\documentclass[lettersize,journal]{IEEEtran}
\usepackage{amsmath,amsfonts}
\usepackage{algorithmic}
\usepackage{algorithm}
\usepackage{array}
\usepackage[caption=false,font=normalsize,labelfont=sf,textfont=sf]{subfig}
\usepackage{textcomp}
\usepackage{stfloats}
\usepackage{url}
\usepackage{verbatim}
\usepackage{graphicx}
\usepackage{cite}
\usepackage{multirow}
\usepackage{colortbl}
\usepackage{xcolor}         % colors
\usepackage{amssymb}
\usepackage{pifont}
\newcommand{\xmark}{\ding{55}}

\usepackage{booktabs}

\usepackage[most]{tcolorbox}
\usepackage[nosfdefault]{comicneue}

\newtcbtheorem[number within=section]{promptbox}{Prompt}%
{colback=gray!10,
 colframe=black!60,
 coltitle=black,
 boxrule=0.2pt,
 arc=2pt,
 left=4pt,right=4pt,top=4pt,bottom=4pt,
 fontupper=\comicneue,
 before upper={\fontsize{6.5pt}{7.5pt}\selectfont}}{pr}

\definecolor{graycell}{gray}{0.9}
\definecolor{teasor_red}{RGB}{255,29,29} 
\definecolor{teasor_blue}{RGB}{0,112,192}
\def\eg{\emph{e.g.}}
\def\ie{\emph{i.e.}}
\def\aka{\emph{aka.}}

\begin{document}

\title{
XOV-Action: Towards Generalizable Open-Vocabulary Action Recognition
}

\author{Kun-Yu Lin, Henghui Ding, Jia-Run Du, Jiaming Zhou, \\
Yi-Xing Peng, Yu-Ming Tang, Zhilin Zhao, Chen Change Loy, Wei-Shi Zheng 
% IEEE Publication Technology,~\IEEEmembership{Staff,~IEEE,}
% <-this % stops a space
\IEEEcompsocitemizethanks{
\IEEEcompsocthanksitem Kun-Yu Lin, Jia-Run Du, Yi-Xing Peng, Yu-Ming Tang and Zhilin Zhao are with the School of Computer Science and Engineering, Sun Yat-sen University, Guangzhou 510275, China. 
E-mail: \{linky5, dujr6, pengyx23, tangym9\}@mail2.sysu.edu.cn, zhaozhlin@mail.sysu.edu.cn. 
\IEEEcompsocthanksitem Henghui Ding is with the Institute of Big Data, Fudan University, Shanghai 200433, China. 
E-mail: hhding@fudan.edu.cn. 
\IEEEcompsocthanksitem Jiaming Zhou is with the AI Thrust, Hong Kong University of Science and Technology (Guangzhou), Guangzhou 511400, China. 
E-mail: jia\_ming\_zhou@outlook.com. 
\IEEEcompsocthanksitem Chen Change Loy is with the College of Computing and Data Science, Nanyang Technological University, Singapore 639798. 
E-mail: ccloy@ntu.edu.sg. 
\IEEEcompsocthanksitem Wei-Shi Zheng is with the School of Computer Science and Engineering, Sun Yat-sen University, Guangzhou 510275, China. 
E-mail: wszheng@ieee.org.
\IEEEcompsocthanksitem Corresponding author: Wei-Shi Zheng. 
This work was partially conducted during Kun-Yu Lin's visit at Nanyang Technological University. 
}
}

\markboth{Journal of \LaTeX\ Class Files,~Vol.~14, No.~8, August~2021}%
{Shell \MakeLowercase{\textit{et al.}}: A Sample Article Using IEEEtran.cls for IEEE Journals}

\IEEEpubid{0000--0000/00\$00.00~\textcopyright~2021 IEEE}

\maketitle

\begin{abstract}
Inspired by the impressive success of image-text foundation models, recent works have proposed to adapt these foundation models to video data, leading to efficient and effective video models for open-vocabulary action recognition. 
However, through a comprehensive evaluation, our work finds that state-of-the-art open-vocabulary action recognition models still struggle with generalization to video domains that they have not encountered. 
To address this limitation, we introduce \textit{generalizable open-vocabulary action recognition}, which aims to develop action recognition models capable of generalizing to both novel action categories and unseen video domains. 
Our work contributes a novel model named XOV-Action to overcome two critical challenges: 
(1) understanding novel action concepts of open-set categories, 
and (2) mitigating the scenario discrepancy between training and test datasets. 
Specifically, XOV-Action first proposes to capture diverse action-related concepts by learning diversified elaboration representations, which enables better generalization to open-set action categories.
Second, XOV-Action learns scene-agnostic video representations to overcome the scene bias, which improves the generalization in unseen video domains.
Additionally, to evaluate models in generalizable open-vocabulary action recognition, we contribute a new cross-domain action benchmark named XOVABench, which covers multiple video domains with varying degrees of gaps and consists of both closed-set and open-set action categories.
Extensive quantitative and qualitative experiments demonstrate that our proposed XOV-Action can effectively improve action recognition performance for both closed-set and open-set categories across video domains. 
The benchmark is available at \url{https://github.com/KunyuLin/XOV-Action/}. 
\end{abstract}


\section{Introduction}
\label{sec:intro}
\IEEEPARstart{A}{ction} recognition aims to recognize what actions humans are performing in videos, which has wide applications in surveillance systems, health monitoring, etc~\cite{DBLP:journals/ijcv/KongF22,9795869}.
Recently, inspired by the impressive success of {image-text foundation models} (\eg, CLIP~\cite{DBLP:conf/icml/RadfordKHRGASAM21}) across various image understanding tasks, pioneer works propose to adapt these models to video data for action recognition~\cite{DBLP:journals/corr/abs-2109-08472,DBLP:conf/eccv/NiPCZMFXL22,DBLP:conf/eccv/JuHZZX22,DBLP:conf/aaai/WuSO23,DBLP:conf/cvpr/RasheedK0KK23,DBLP:conf/icml/WengYLWJ23}.
Different from traditional action models that focus on closed-set recognition~\cite{DBLP:conf/nips/SimonyanZ14,DBLP:conf/eccv/WangXW0LTG16,DBLP:conf/iccv/LinGH19,DBLP:conf/cvpr/CarreiraZ17,DBLP:conf/icml/BertasiusWT21},
this new {image-text foundation-based} paradigm leads to efficient video learners with remarkable \textit{open-vocabulary action recognition} abilities, \ie, they achieve state-of-the-art performance for \textit{both closed-set and open-set action categories}\footnotemark[1] on various video datasets with moderate training cost.
\IEEEpubidadjcol
Such open-vocabulary abilities significantly improve the practical utility of action recognition models.
\footnotetext[1]{Closed-set categories refer to the categories seen during training, while open-set categories refer to the others.}

\begin{figure}[t]
\centering
\includegraphics[width=0.95\linewidth]{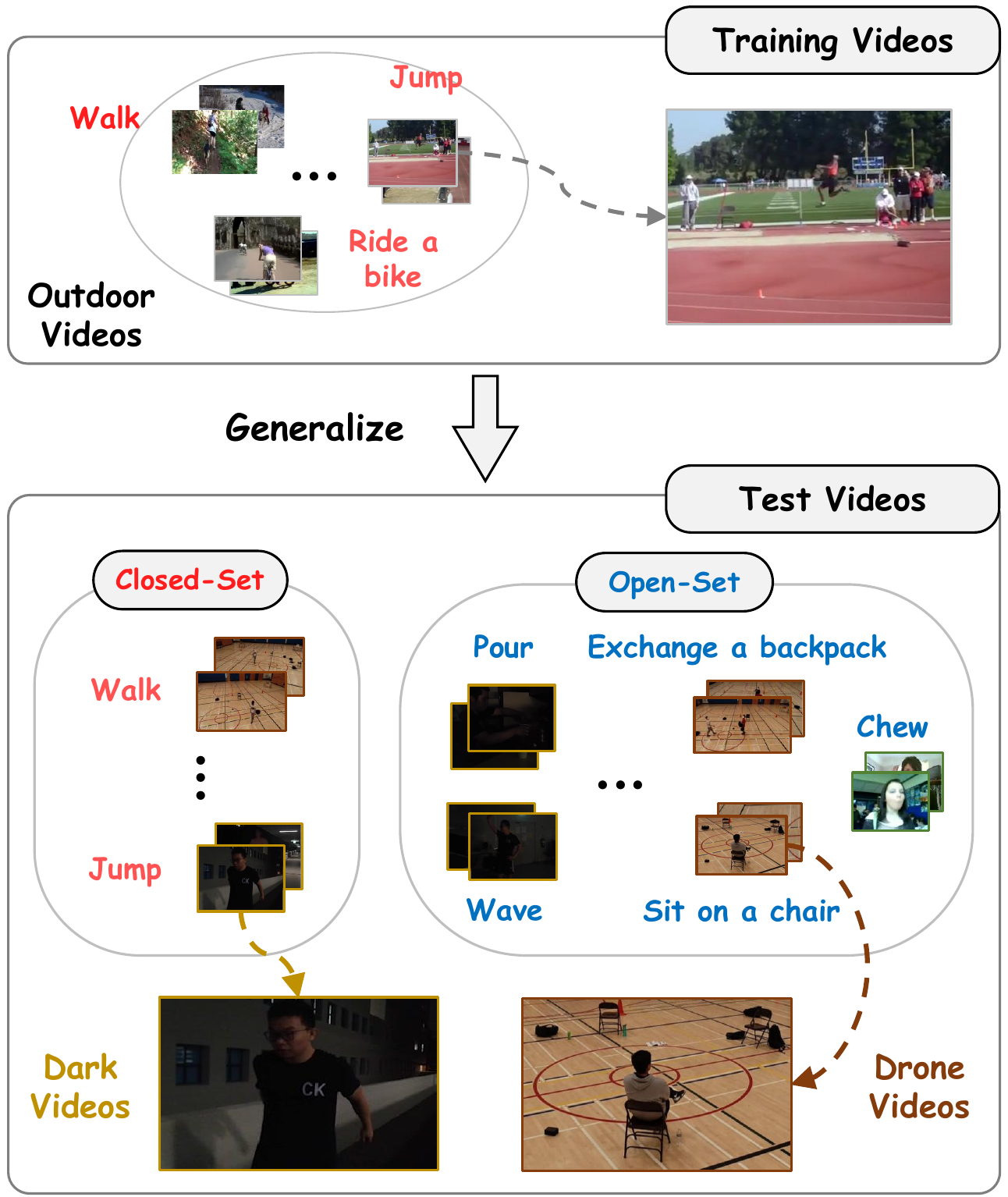}
\vskip -0.05in
\caption{
The demonstration of our proposed \textit{generalizable open-vocabulary action recognition} task, which aims to develop models capable of recognizing \textit{both close-set and open-set categories across video domains}, \eg, from a outdoor video domain to a dark video domain. 
\textcolor{teasor_red}{Red texts} denote the closed-set categories, and \textcolor{teasor_blue}{blue texts} denote the open-set categories.  
Best viewed in color.
}
\label{fig:intro}
\vskip -0.1in
\end{figure}

In this work, we study an underexplored task termed \textit{generalizable open-vocabulary action recognition}, where training and test videos are drawn from different domains and may encompass non-overlapping action categories. 
As shown in Figure~\ref{fig:intro}, the goal of this task is to develop open-vocabulary action recognition models capable of \textit{generalizing to test domains not encountered during training}. 
Such generalization abilities are crucial for action recognition models, since models often suffer from environment, viewpoint, and sensor changes when deployed in real-world applications~\cite{DBLP:conf/iccv/ChenKAYCZ19,DBLP:journals/pami/YaoWWYL22}. 
For example, surveillance systems will encounter actions performed under illumination shifts caused by day-night change or weather change. 
Therefore, we expect that action recognition models with open-vocabulary abilities can robustly address video domain shifts.

\begin{figure}[t]
\centering
\includegraphics[width=0.98\linewidth]{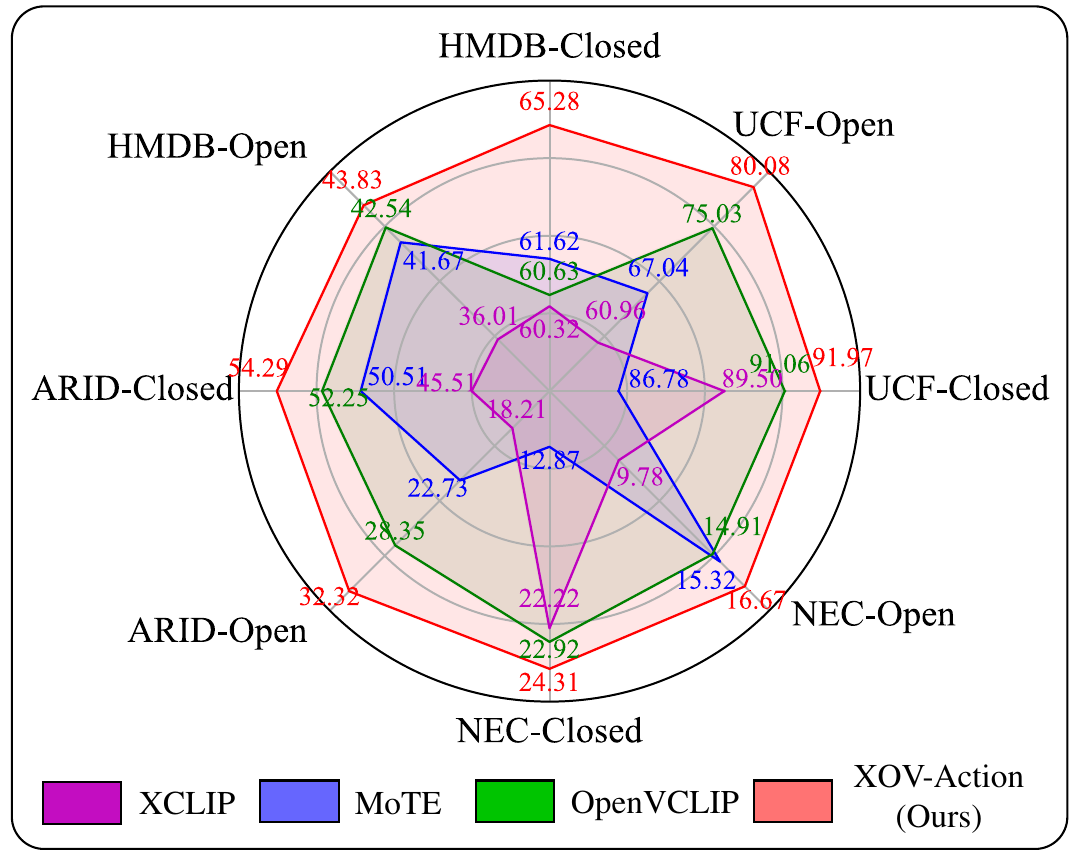}
\vskip -0.1in
\caption{
We conduct a evaluation for state-of-the-art open-vocabulary action recognition models on four test datasets, namely UCF~\cite{DBLP:journals/corr/abs-1212-0402}, HMDB~\cite{DBLP:conf/iccv/KuehneJGPS11}, ARID~\cite{xu2020arid} and NEC-Dr~\cite{DBLP:conf/wacv/ChoiSCH20}.
These four test datasets exhibiting various levels of domain gaps in comparison to the training dataset, \ie, UCF has a small gap, HMDB has a moderate gap, ARID and NEC-Dr have large domain gaps.
For each test dataset, we report the accuracy of closed-set and open-set action categories, which are identified according to the training categories in Kinetics400~\cite{DBLP:conf/cvpr/CarreiraZ17}.
As shown in the figure, previous state-of-the-art open-vocabulary models exhibit limited performance when recognizing actions in \textit{unseen} test domains. 
Please refer to {Table~\ref{tab:k400}} for the full results.
Best viewed in color.
}
\label{fig:intro_sota}
% \vskip -0.1in
\end{figure}

By conducting a comprehensive evaluation on state-of-the-art open-vocabulary action recognition models, our work reveals that, despite their remarkable success, state-of-the-art models still struggle with generalization to \textit{unseen} video domains. 
As shown in {Figure~\ref{fig:intro_sota} and Table~\ref{tab:k400}}, the open-set recognition performance of previous open-vocabulary models~\cite{DBLP:conf/eccv/NiPCZMFXL22,DBLP:conf/icml/WengYLWJ23,DBLP:nips/abs-2410-10589} are far from reaching saturation even in domains with \textit{moderate} domain gaps, \eg, the best open-set accuracy on HMDB is only {42.54\%}. 
In addition, our evaluation shows that previous open-vocabulary models exhibit significantly degraded performance on closed-set categories when tested in domains with \textit{large} domain gaps. 
For example, even the top-performing Kinetics400-trained model achieves merely {52.74\%} closed-set accuracy on dark videos in ARID, due to the large scenario gap between the Kinetics and ARID domains.
Overall, generalizable open-vocabulary action recognition is a challenging task, since it requires recognizing both closed-set and open-set action categories in unseen video domains.

In principle, there are two critical challenges in the \textit{generalizable} open-vocabulary action recognition task, namely \textit{novel action concepts of open-set categories} and \textit{scenario discrepancy between training and test videos}:

1) The first challenge lies in understanding the novel action concepts of open-set categories, which is fundamental to category generalization. 
Specifically, previous open-vocabulary models~\cite{DBLP:journals/corr/abs-2109-08472,DBLP:conf/cvpr/RasheedK0KK23,DBLP:conf/icml/WengYLWJ23} strongly rely on the text representations of category names to recognize open-set action categories. 
However, action concepts encoded by these simple name texts of open-set categories are usually \textit{vague and unfamiliar} to models, since \textit{video samples of open-set categories are not accessible for model training}.  
For example, it is challenging for a recognition model to distinguish the action ``long jump'' from the action ``high jump'' solely based on name texts, if the model do not see any videos of these two categories during training.
This is because these two name texts are similar as both categories share the word ``jump'', and a model can hardly understand the meaning of ``long'' and ``high'' in the context of the respective actions \textit{without seeing any video samples}.

2)
Secondly, in generalizable open-vocabulary action recognition, domain gap is a major obstacle to model generalization. 
It is an important challenge widely discussed in traditional action recognition works~\cite{DBLP:conf/cvpr/SultaniS14,DBLP:conf/iccv/ChenKAYCZ19}, but remains \textit{underexplored} in open-vocabulary works. 
To bridge the domain gap, we observe that two videos from different domains usually have distinct scenarios, and this can easily cause scene bias during model training. 
Specifically, the scene bias arises from the strong associations between actions and specific scenarios in training videos (\aka, spurious correlation~\cite{DBLP:conf/eccv/LiLV18,DBLP:conf/nips/ChoiGMH19,DBLP:conf/eccv/ChoiSSH20,DBLP:conf/cvpr/WangGLLM0PHJS21,DBLP:conf/iccv/LiLZL23,DBLP:conf/iccv/0001LWWZDY023}), and it would hinder model generalization across domains due to the scenario differences. 
For example, as shown in Figure~\ref{fig:intro}, if humans always perform jumping on track-and-field grounds in the training domain, the trained models are prone to recognize this action by the track-and-field grounds, since static scenes are much easier to fit~\cite{DBLP:conf/eccv/LiLV18,DBLP:conf/nips/ChoiGMH19,DBLP:conf/cvpr/WangGLLM0PHJS21}. 
However, humans may perform jumping in hallways in test domains, thus recognizing ``jump'' based on track-and-field grounds would result in recognition errors across domains.

In this work, we propose a novel model, named XOV-Action, to overcome the above two challenges for generalizable open-vocabulary action recognition.  
\underline{Firstly}, to boost the understanding of novel action concepts for open-set categories, our XOV-Action proposes to capture diverse action-related concepts from videos by Diversified Elaboration Representation Learning. 
By elaborating action concepts using multiple textual descriptions, XOV-Action learns diverse concepts during training and associates them with open-set categories during testing, thereby improving the recognition of open-set action categories.
\underline{Secondly}, to mitigate the scenario discrepancy, our XOV-Action proposes to learn scene-agnostic video representations by Scene-Aware Video-text Alignment. 
By introducing scene-encoded text prompts, XOV-Action distinguishes video representations apart from scene-encoded text representations, which encourages the video encoder to downweight the attention on scene information and thus pay more attention to action information, thereby improving the generalization across domains. 

In addition, to evaluate models for generalizable open-vocabulary action recognition, we establish a new cross-domain action benchmark named XOVABench, which covers multiple video domains with varying degrees of gaps and consists of both closed-set and open-set action categories.
Extensive quantitative and qualitative experiments demonstrate that our proposed XOV-Action can effectively improve the action recognition performance for both closed-set and open-set categories across domains.

In summary, our contributions are listed as follows:

(1) Our proposed model, named XOV-Action, can capture rich action-related concepts from videos by Diversified Elaboration Representation Learning. 
By leveraging textual descriptions of action categories, XOV-Action learns diverse concepts during training and associates them with open-set categories during test.  
Extensive experiments demonstrate the effectiveness of our XOV-Action in recognizing novel actions across domains.

(2) Our proposed XOV-Action also learns scene-agnostic video representations by Scene-Aware Video-text Alignment, encouraging the video encoder to downweight the attention on scene information and thus pay more attention to inherent action information. 
Extensive experiments demonstrate the effectiveness of our XOV-Action in bridging domain gaps.

(3) We establish a CROSS-domain Open-Vocabulary Action recognition Benchmark, dubbed XOVABench, which consists of four test domains exhibiting various levels of domain gaps in comparison to the training domains.  
We identify closed-set and open-set categories for each test domain, and thus provide a comprehensive way to evaluate open-vocabulary action models across various situations.

\section{Related Work}
\label{sec:related_work}

\subsection{Action Recognition} 
Action recognition aims to recognize human actions in videos, which has broad applications in real-world~\cite{DBLP:journals/ijcv/KongF22,9795869,DBLP:conf/eccv/LiHWZMZ24}.
In the past decade, motivated by the success of deep learning~\cite{DBLP:conf/nips/KrizhevskySH12,DBLP:journals/corr/SimonyanZ14a,DBLP:conf/cvpr/SzegedyLJSRAEVR15,DBLP:conf/cvpr/HeZRS16,DBLP:conf/nips/SutskeverVL14,DBLP:conf/emnlp/ChoMGBBSB14,DBLP:conf/nips/VaswaniSPUJGKP17}, many video classification architectures have been proposed.
These architectures can be primarily categorized into 2D CNNs, 3D CNNs and Video Transformers.
Typically, 2D CNNs adopt 2D convolution for spatial modeling, and conduct temporal modeling beyond spatial modeling~\cite{DBLP:conf/eccv/WangXW0LTG16,DBLP:conf/eccv/ZhouAOT18,DBLP:conf/cvpr/ZhouLLZ21,DBLP:journals/corr/abs-2503-22405} or embed temporal shift into spatial modeling~\cite{DBLP:conf/iccv/LinGH19,DBLP:conf/aaai/ShaoQL20,DBLP:conf/cvpr/SudhakaranEL20}.
3D CNNs extends 2D convolution to 3D convolution for adapting video data~\cite{DBLP:conf/iccv/TranBFTP15,DBLP:conf/cvpr/CarreiraZ17,DBLP:conf/cvpr/TranWTRLP18,DBLP:conf/iccv/TranWFT19,DBLP:conf/cvpr/Feichtenhofer20,DBLP:conf/iclr/ZhangGHS020,DBLP:conf/iclr/LiLWWQ21}. 
By adopting attention mechanisms, Video Transformers expand the receptive field of 2D and 3D CNNs, leading to remarkable performance~\cite{DBLP:conf/cvpr/GirdharCDZ19,DBLP:journals/corr/abs-2102-00719,DBLP:conf/mm/ZhangHN21,DBLP:conf/iccv/ZhangLLSZBCMT21,DBLP:conf/icml/BertasiusWT21,DBLP:conf/iccv/Arnab0H0LS21,DBLP:journals/tmm/ZhouLQZ24,DBLP:conf/iccv/WangLDM023}. Although above models show promising performance for closed-set action recognition, they usually lack the ability to recognize open-set categories.

Recently, inspired by the success of {image-text foundation models} (especially CLIP~\cite{DBLP:conf/icml/RadfordKHRGASAM21}), some pioneer works propose adapting these models to video data for action recognition~\cite{DBLP:conf/eccv/LinGZGMWDQL22,DBLP:journals/corr/abs-2109-08472,DBLP:conf/eccv/NiPCZMFXL22,DBLP:conf/eccv/LinGZGMWDQL22,DBLP:conf/eccv/JuHZZX22,DBLP:conf/aaai/WuSO23,DBLP:conf/cvpr/RasheedK0KK23,DBLP:conf/icml/WengYLWJ23,DBLP:conf/cvpr/WuWLWYO23,DBLP:conf/iccv/QingZHZGZS23,DBLP:journals/corr/abs-2304-02560,DBLP:conf/iccv/ChenCLLP23,DBLP:journals/corr/abs-2402-03241,DBLP:conf/cvpr/LiuHLFWL23,DBLP:conf/mm/WangDYD23,DBLP:conf/aaai/WangXJ0MZD0024,DBLP:conf/cvpr/ChenCLZX024,DBLP:conf/cvpr/ZhangWLTSY24,DBLP:conf/eccv/KimHKH24,DBLP:nips/abs-2410-10589}.
Owing to the image-text alignment power of image-text foundation models, video learners are endowed with remarkable open-vocabulary action recognition abilities with moderate training cost.
These methods adapt image-text foundation models to video data in various ways.
For example, ActionCLIP~\cite{DBLP:journals/corr/abs-2109-08472} stacks temporal fusion layers on top of the image encoder for modeling temporal dynamics, Ju et al.~\cite{DBLP:conf/eccv/JuHZZX22} cooperate continuous prompting with temporal Transformer, and X-CLIP~\cite{DBLP:conf/eccv/NiPCZMFXL22} proposes cross-frame communication attention for temporal modeling.
Different from other works, two pioneer works OpenVCLIP~\cite{DBLP:conf/icml/WengYLWJ23} and FROSTER~\cite{DBLP:journals/corr/abs-2402-03241} propose to improve open-set generalization by harnessing the power of raw CLIP.
In our work, we focus on the underexplored \textit{cross-domain} scenarios of open-vocabulary action recognition, which aims to develop generalizable open-vocabulary action recognition models for \textit{unseen} video domains.

\subsection{Generalizable Action Recognition} 
Generalizable action recognition studies the \textit{generalization capabilities} of action recognition models. 
In this field, \textit{zero-shot action recognition}~\cite{DBLP:conf/cvpr/LiuKS11} aims to recognize novel categories of actions not encountered during training. 
Existing works typically learn alignment models between the visual space of videos and the semantic space of
class descriptions~\cite{DBLP:conf/iccv/KodirovXFG15,DBLP:conf/iccv/JainGMS15,DBLP:conf/aaai/GanLYMH16,DBLP:conf/eccv/FuHXFG14,DBLP:journals/ijcv/WangC17,DBLP:conf/iccv/ChenH21,DBLP:journals/corr/abs-2401-11654}, and this zero-shot task establishes a solid foundation for open-vocabulary action recognition. 
Additionally, \textit{cross-domain action recognition} aims to learn video classification models by transferring knowledge from source domains to target domains. 
This mainly includes two tasks, namely \textit{domain-adaptive action recognition} and \textit{domain-generalizable action recognition}.
In domain-adaptive action recognition~\cite{DBLP:conf/cvpr/SultaniS14,DBLP:journals/ivc/XuZWF16,DBLP:conf/bmvc/JamalNDV18,DBLP:conf/iccv/ChenKAYCZ19,DBLP:conf/iccv/XuYCCLM21,DBLP:journals/pami/BustoIG20,DBLP:conf/eccv/XuYCWWC22,DBLP:journals/pami/LinZZ25}, unlabeled videos from target domains are accessible for training, thus prevailing works usually focus on developing models oriented to specific target domains. 
Typical works address domain-adaptive action recognition by learning cross-domain invariance~\cite{DBLP:conf/iccv/ChenKAYCZ19,DBLP:conf/aaai/PanCAN20,xu2022aligning,DBLP:conf/mm/LuoHW0B20,DBLP:conf/nips/SahooSPSD21}, and other works usually explore leveraging the multi-modal nature of video data~\cite{DBLP:conf/cvpr/MunroD20,DBLP:conf/cvpr/SongZ0Y0HC21,DBLP:conf/iccv/KimTZ0SSC21,DBLP:conf/cvpr/YangHSS22,Zhang_2022_CVPR}.
Differently, domain-generalizable action recognition~\cite{DBLP:journals/pami/YaoWWYL22,DBLP:conf/wacv/PlanamentePAC22,DBLP:conf/nips/LinDGZ023,DBLP:journals/corr/abs-2306-08713} aims to learn models generalizable in \textit{unseen} test domains, where videos of target domains are not accessible during training. 
To addess this task in the absence of target videos, Yao et al. assume that local features are more invariant across domains~\cite{DBLP:journals/pami/YaoWWYL22}, and Lin et al. propose to learn diverse action features~\cite{DBLP:conf/nips/LinDGZ023}.
Our generalizable open-vocabulary action recognition task is closely related to domain-generalizable action recognition, as test domains remain unseen during training.
However, our task is significantly more challenging as we strive for open-vocabulary abilities, leading to significantly different technical designs.

\subsection{Vision-Language Pretraining and Its Applications}
In recent years, research on image-text foundation models has made great progress~\cite{DBLP:conf/icml/RadfordKHRGASAM21,DBLP:conf/nips/LiSGJXH21,DBLP:conf/cvpr/Li0HFH23,DBLP:conf/nips/AlayracDLMBHLMM22,DBLP:journals/corr/abs-2111-11432,DBLP:journals/tmlr/YuWVYSW22,DBLP:conf/ijcai/DuLLZ22,DBLP:conf/iccv/ZhaiM0B23,DBLP:journals/tmlr/OquabDMVSKFHMEA24,DBLP:conf/iccv/KirillovMRMRGXW23}, \eg, CLIP~\cite{DBLP:conf/icml/RadfordKHRGASAM21}, SigLIP~\cite{DBLP:conf/iccv/ZhaiM0B23}, CoCa~\cite{DBLP:journals/tmlr/YuWVYSW22}. 
Among these works, CLIP (Contrastive Language-Image Pretraining)~\cite{DBLP:conf/icml/RadfordKHRGASAM21} is one of the most representative, and it is the foundation of many open-vocabulary works. 
By utilizing web-scale paired image-text data for training, CLIP shows robust zero-shot object recognition abilities.
Also, many advanced works have demonstrated that CLIP can effectively solve downstream tasks by efficient adaptation~\cite{DBLP:journals/ijcv/ZhouYLL22,DBLP:conf/cvpr/0012LJ0W23,DBLP:conf/nips/PanLZS022,DBLP:conf/cvpr/ZhouYL022,DBLP:conf/cvpr/RasheedK0KK23,DBLP:conf/cvpr/LiuHLFWL23}.
Moreover, integrating CLIP with specialized techniques shows remarkable open-vocabulary abilities on various image understanding tasks, \eg, object detection and segmentation~\cite{DBLP:journals/corr/abs-2306-15880,DBLP:conf/iclr/GuLKC22,DBLP:conf/iclr/LiWBKR22,DBLP:conf/eccv/LinGZGMWDQL22,DBLP:conf/cvpr/LLMDET}.
Although large-scale image-text pretraining has achieved great success, video-text pretraining still has room for development~\cite{DBLP:conf/cvpr/WasimNKKS23,DBLP:conf/emnlp/XuG0OAMZF21,DBLP:conf/cvpr/Li0LNH22,DBLP:conf/nips/Wang0WLZZX0JY22,DBLP:journals/corr/abs-2304-06708,DBLP:conf/cvpr/HuangLFWSJ23,DBLP:conf/cvpr/ChengW0CBB23,DBLP:journals/corr/abs-2403-14870,DBLP:conf/aaai/MaJWHZ024,DBLP:journals/corr/abs-2502-13923,DBLP:conf/eccv/WangLLYHCPZWSJLXZHQWW24,DBLP:journals/corr/abs-2405-03770}.
It is because videos are inherently more complex than images, and large-scale paired video-text datasets are less available.
Therefore, it is valuable to develop methods for adapting pretrained image-text models to video understanding.

\begin{figure*}[t]
\vskip -0.1in
\centering
\includegraphics[width=1.0\linewidth]{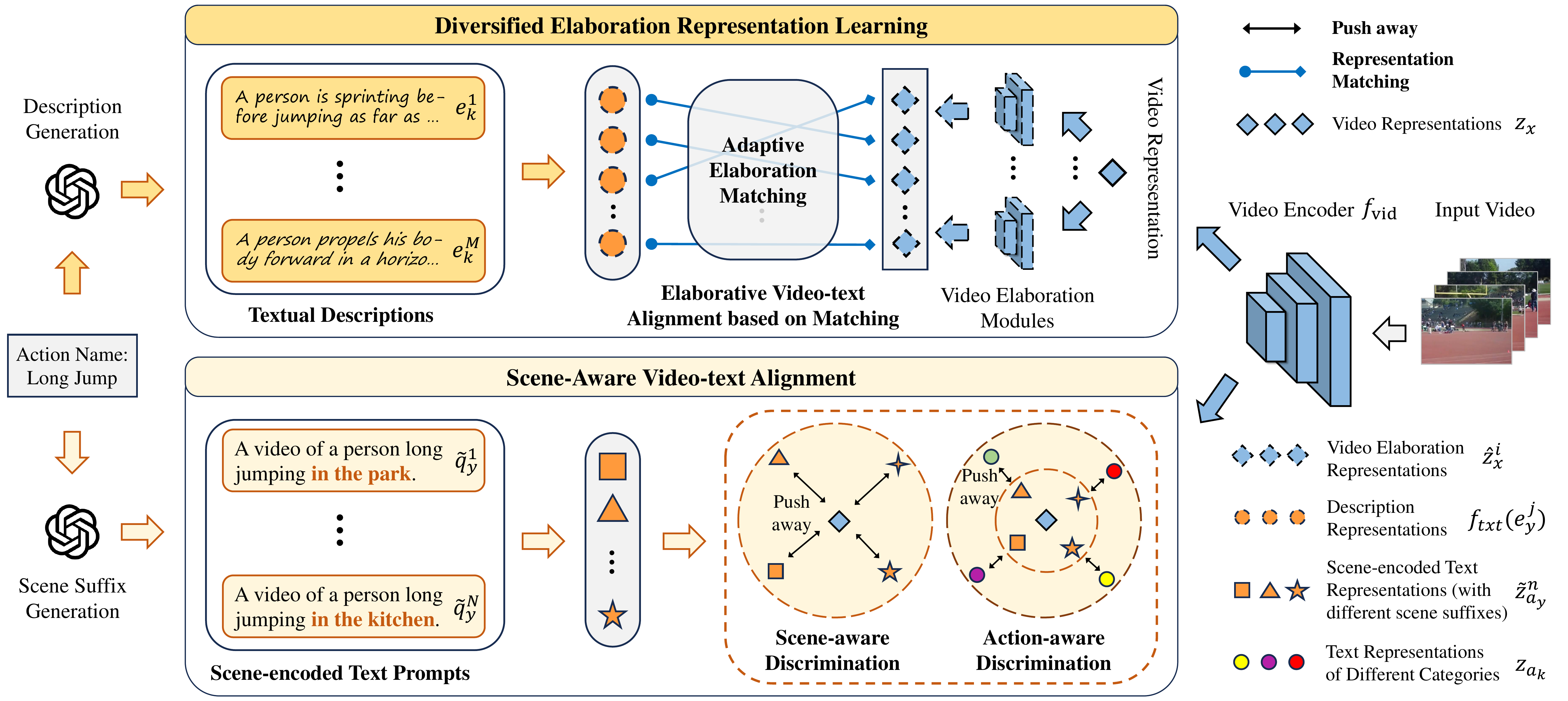}
\vskip -0.1in
\caption{
An overview of our proposed XOV-Action model, which aims to overcome two critical challenges of the generalizable open-vocabulary action recognition task. 
First, our XOV-Action proposes Diversified Elaboration Representation Learning to boost the understanding of novel action concepts for open-set categories. 
By leveraging the Elaborative Video-text Alignment loss with Adaptive Elaboration Matching, XOV-Action captures diverse action-related concepts under the guidance of multiple textual descriptions. 
Second, to defend against the scene bias, our XOV-Action proposes Scene-Aware Video-text Alignment to learn scene-agnostic video representations. 
By leveraging the Scene-aware Discrimination and Action-aware Discrimination losses, XOV-Action encourages the video encoder to downweight the attention on scene information under the guidance of scene-encoded text prompts.
Best viewed in color.
}
\label{fig:model}
\vskip -0.05in
\end{figure*}

\section{The Proposed XOV-Action Model}

\subsection{Problem Formulation}
This work focuses on the generalizable open-vocabulary action recognition task.
In this task, a set of labeled videos $\mathcal{D}=\{(x_i, y_i)\}_{i=1}^{N_s}$ from a source domain are given for training, where $x_i$ denotes a source video, $y_i$ denotes its ground-truth action label index and $N_s$ denotes the number of videos.
The source domain consists of $K$ action categories, $y_i\in\{1, 2, \dots, K\}$, and the action name texts are denoted by $\{a_1, a_2, \dots, a_K\}$.
For notational simplicity, we omit the sample index $i$ of $x_i$ and $y_i$ in the following formulations when it does not cause ambiguity.
Given only source videos for training, our goal is to develop a model that is \textit{generalizable} in \textit{unseen} target domains, where the source and target domains follow different data distributions and different label spaces. 
Following the standard protocol of previous open-vocabulary action recognition works~\cite{DBLP:conf/cvpr/RasheedK0KK23}, we sample $T$ frames from each video as model input during training and test.

\subsection{Model Overview}
Our XOV-Action model is built based on CLIP~\cite{DBLP:conf/icml/RadfordKHRGASAM21} composed of a video encoder $f_{\text{vid}}(\cdot)$ and a text encoder $f_{\text{txt}}(\cdot)$, following protocols of previous open-vocabulary action recognition works~\cite{DBLP:conf/cvpr/RasheedK0KK23,DBLP:conf/icml/WengYLWJ23}. 
Specifically, we construct a video encoder by integrating patches from neighboring frames in each self-attention layer of the original CLIP image encoder.
For the video $x$, the global video representation $z_{x} = f_{\text{vid}}(x)$ is obtained by the average of local video representations, where each local representation corresponds to the representation of one frame.
We use the original CLIP text encoder and keep it frozen during training. 
Following previous works, we use a video-text alignment loss to adapt video data, which is formulated as follows:
\begin{equation}
\begin{aligned}
    L_{\text{vta}} = - \log{\frac{\exp{( s(z_{x}, z_{a_y}) / \tau )}}{\sum_{k=1}^{K} \exp{( s(z_{x}, z_{a_k}) / \tau )}}},
    \label{eq:baseline}
\end{aligned}
\end{equation}
where $s(\cdot, \cdot)$ denotes the cosine similarity, $\tau$ is the temperature, $z_{a_k}=f_{\text{txt}}(g(a_k))$ is the text representation of the $k$-th action name text.
The function $g(\cdot)$ converts an action name into a text prompt in the form of ``\texttt{a video of a person [doing something].}'', \eg, ``\texttt{a video of a person abseiling.}'' for the action ``abseiling''.
The above loss follows the standard formulation of previous contrastive-learning-based works, \eg, MoCo~\cite{DBLP:conf/cvpr/He0WXG20} and CLIP~\cite{DBLP:conf/icml/RadfordKHRGASAM21}.

An overview of our proposed XOV-Action is shown in Figure~\ref{fig:model}.
Our proposed XOV-Action includes two key components as technical contributions, namely \textit{Diversified Elaboration Representation Learning} and \textit{Scene-Aware Video-text Alignment}. 
Firstly, XOV-Action proposes Diversified Elaboration Representation Learning to capture rich action-related concepts, by elaborating action concepts using multiple textual descriptions.
Through an Elaborative Video-text Alignment loss with Adaptive Elaboration Matching, our model learns diverse elaboration representations for each video, which boosts the understanding of novel action concepts for open-set categories.
Secondly, by introducing scene-encoded text prompts, XOV-Action proposes a Scene-Aware Video-text Alignment method, which consists of a Scene-aware Discrimination loss and an Action-aware Discrimination loss. 
Accordingly, our model can learn scene-agnostic video representations to defend against the scene bias and improve the generalization in unseen domains. 
In what follows, we illustrate XOV-Action in detail.

\subsection{Diversified Elaboration Representation Learning}
Our proposed XOV-Action model first focuses on a fundamental challenge of generalizable open-vocabulary action recognition models, namely the understanding of novel action concepts for open-set categories.
Specifically, we propose a novel Diversified Elaboration Representation Learning method, which introduces multiple textual descriptions as auxiliary supervision in training. 
By incorporating the proposed Elaborative Video-text Alignment loss with Adaptive Elaboration Matching, our model learns multiple elaboration representations for each video, which encode rich concept information related to action categories. 
After learning action-related concepts, our model can associate concepts with open-set categories during testing through action descriptions, thereby improving the recognition of open-set action categories.  

First of all, we introduce a set of $M$ textual descriptions for the $k$-th action category, denoted by $\{e^1_k, e^2_k, \dots, e^M_k\}$. 
Specifically, for each action category, we ask GPT-4~\cite{openai2023gpt4} to automatically generate a diverse set of textual descriptions based on the category name, which does not involve any human annotation cost. 
Compared with simple name texts, these textual descriptions involve much more details and concept information about the corresponding action category, which helps distinguish different action categories. 
For example, when distinguishing open-set actions ``long jump'' and ``high jump'', it is difficult to understand the meaning of ``long'' and ``high'' in the context of the respective actions \textit{without seeing any video samples during training}, since ``long jump'' and ``high jump'' are novel action concepts to models. 
After introducing the textual description ``\texttt{A person is seen accelerating on a track, taking a leap, and landing in a distant sand pit.}'' for the action ``long jump'' and the textual description ``\texttt{A person attempts to jump over a horizontal bar at the greatest height possible, using a specific technique.}'' for the action ``high jump'', it is much easier for models to distinguish these two actions.
This is because these two textual descriptions contain more action-related concept information, \eg, ``\texttt{sand pit}'' and ``\texttt{horizontal bar}'', and thus models can leverage these concepts to distinguish actions ``long jump'' and ``high jump''.

\textbf{Adaptive Elaboration Matching:}
By leveraging these textual descriptions as guidance, we propose to learn rich action-related concepts by Diversified Elaboration Representation Learning.  
Specifically, our model learns $C$ elaboration representations for each video, denoted by $\{\hat{z}_x^{1}, \hat{z}_x^{2}, \dots, \hat{z}_x^{C}\}$. 
The $i$-th elaboration representation $\hat{z}_x^{i}$ is produced by the $i$-th video elaboration module, \ie, $\hat{z}_x^{i} = h_i(z_x)$, and there are $C$ lightweight video elaboration modules on top of our vision encoder $f_{\text{vid}}(\cdot)$.
During training, the learning of an elaboration representation is supervised by its best-matched textual description, since the introduced descriptions for each action category form an unordered set, \ie, their ordering carries no semantic meaning.
We formulate an adaptive representation matching problem to find the best-matched descriptions, termed Adaptive Elaboration Matching.  
Suppose that the number of elaboration representations $C$ is larger than the number of textual descriptions $M$.
Then, for a video of the $k$-th action category, our Adaptive Elaboration Matching between elaboration representations and textual descriptions is formulated as follows:
\begin{equation}
\begin{aligned}
    O^k = & \mathop{\arg\max}\limits_{O^k\in\{0,1\}^{C\times M}} \sum_{i=1}^C \sum_{j=1}^M s(\hat{z}_x^{i}, f_{\text{txt}}(e^j_k)) \cdot O^k_{i,j}, \\
    \text{s.t.}\quad 
    & \sum\nolimits_{j=1}^M O^k_{i,j} \le 1,\ \forall i,~\sum\nolimits_{i=1}^C O^k_{i,j} = 1,\ \forall j,
    \label{eq:matching}
\end{aligned}
\end{equation}
where $f_{\text{txt}}(e^j_k)$ is the text representation of description $e^j_k$. 
In this formulation, $O^k\in\{0,1\}^{C\times M}$ is the assignment matrix composed of binary elements, and the superscript $k$ denotes the $k$-th category. 
The matrix element $O^k_{i,j}\in\{0, 1\}$ indicates the matching between the elaboration representation $\hat{z}_x^{i}$ and description $e^j_k$, \ie, $O^k_{i,j}=1$ indicates that $\hat{z}_x^{i}$ is matched with $e^j_k$ and $O^k_{i,j}=0$ indicates no match. 
The constraints $\sum_{j=1}^M O^k_{i,j} \leq 1$ and $\sum_{i=1}^C O^k_{i,j} = 1$ are introduced to ensure an injective one-to-one matching.
The matching problem presented in Eq.~\eqref{eq:matching} is solved by the Hungarian algorithm~\cite{DBLP:books/daglib/p/Kuhn10}. 
Overall, Adaptive Elaboration Matching avoids arbitrary alignments between the unordered sets of video elaboration representations and textual descriptions. 
This reduces supervision noise and facilitates the learning of diverse elaboration representations that encode rich action-related concepts.

\textbf{Elaborative Video-text Alignment:}
In addition, since the textual descriptions are generated by GPT-4, a single video may not contain all details mentioned by the descriptions of its ground-truth category. 
Thus, simply using all descriptions for each video may introduce noise during training. 
Accordingly, for each video, we propose a confidence-aware elaboration selection strategy to select the top-$\hat{C}$ most relevant descriptions from every action category during training. 
Formally, we propose an Elaborative Video-text Alignment loss for learning diversified elaboration representations, which is given as follows:
\begin{equation}
\begin{aligned}
    L_{\text{eva}} = - \log{\frac{\exp{\left( \frac{1}{\tau\cdot\hat{C}} \sum_{(i,j)\in \mathcal{O}^y_{\hat{C}}} s(\hat{z}_x^{i}, f_{\text{txt}}(e^j_y)) \right)}}{\sum_{k=1}^{K} \exp{\left( \frac{1}{\tau\cdot\hat{C}} \sum_{(i,j)\in \mathcal{O}^k_{\hat{C}}} s(\hat{z}_x^{i}, f_{\text{txt}}(e^j_k)) \right)}}},
    \label{eq:elabo}
\end{aligned}
\end{equation}
where $\mathcal{O}^k_{\hat{C}}$ is a subset of matches extracted from $O^k$, consisting of the index pairs $(i,j)$ that yield the top-$\hat{C}$ highest similarity $s(\hat{z}_x^{i}, f_{\text{txt}}(e^j_k))$.
This formulation enables the model to selectively align each video sample with the top-$\hat{C}$ best-matched textual descriptions of its ground-truth category, while contrasting the video sample against descriptions from other categories.
Guided by this loss, our model can learn diverse elaboration representations that encode rich and accurate action-related concepts under the supervision of diverse textual descriptions.
During testing, we also introduce multiple textual representations for each action category in target domains. 
As a result, our model can associate learned concepts with the novel action concepts of open-set categories, thus improving
the recognition of open-set action categories. 

Notably, our proposed Diversified Elaboration Representation Learning is robust to different LLM-generated textual descriptions, as shown in Tables~A1-A5 in the Appendix. 
In practical deployment, these category-level descriptions introduce a lightweight LLM dependency, requiring a one-time LLM generation step before training or inference.
While current LLMs work reliably for common actions, they may produce overly generic or imprecise descriptions for rare actions, \eg, niche cultural practices.
In such cases, we suggest a lightweight human-in-the-loop check before deployment.

\subsection{Scene-Aware Video-text Alignment}
More importantly, in the generalizable open-vocabulary action recognition task, domain gap is a major obstacle to model generalization, and it is largely ignored by existing open-vocabulary works. 
Therefore, to bridge the domain gap and improve the generalization in unseen video domains, we focus on the scenario difference, a key difference between two video domains.
Accordingly, to mitigate the scenario discrepancy across video domains, we propose a novel Scene-Aware Video-text Alignment method to learn scene-agnostic video representations.
The key idea of our Scene-Aware Video-text Alignment is to distinguish video representations apart from scene-encoded text representations, which encourages the video encoder to downweight the attention on scene information in videos.

\textbf{Scene-aware Discrimination:}
First of all, we randomly sample $N$ scene suffixes and construct scene-encoded text prompts for each training video.
Each scene suffix is in the form of ``\texttt{[at/on/in the/a scene]}'', \eg, ``\texttt{in the park}'', ``\texttt{on the street}''.
In our implementation, we ask GPT-4~\cite{openai2023gpt4} to automatically generate a pool of scene suffixes for random sampling, which involves no human annotation cost.
Based on these suffixes, we construct scene-encoded text prompts of ground-truth action category for each video, which is in the form of ``\texttt{a video of a person [doing something] [at/on/in the/a scene].}''. 
For example, for the action ``abseiling'', we construct $N$ scene-encoded text prompts, \eg, ``\texttt{a video of a person abseiling in the park.}''.

Then, based on the scene-encoded text prompts, we design a Scene-aware Discrimination loss, which is formulated as follows:
\begin{equation}
\begin{aligned}
    L_{\text{scene}} = - \log{\frac{\exp{( s(z_{x}, z_{a_y}) )}}{ \exp{( s(z_{x}, z_{a_y}) )} + \sum_{n=1}^{N} \exp{( s(z_{x}, \tilde{z}^n_{a_y}) )}}}.
    \label{eq:push}
\end{aligned}
\end{equation}
In this loss, $\tilde{z}^n_{a_y}=f_{\text{txt}}(\tilde{q}^n_y)$ is the representation of a scene-encoded text prompt, which encodes the semantic information of the $n$-th scene.
We denote the $n$-th scene-encoded text prompt by $\tilde{q}^n_y=\tilde{g}(a_y, n)$, where the function $\tilde{g}(\cdot, n)$ transforms an action name into a scene-encoded text prompt by incorporating the $n$-th scene suffix. 
According to Eq.~\eqref{eq:push}, our proposed Scene-aware Discrimination loss pushes the video representations away from the scene-encoded text representations in video-text alignment.
In this way, this loss encourages the video encoder to pay less attention to scene information and thus pay more attention to action information, by leveraging the strong power of CLIP text encoder. 
As a result, we can mitigate the scene bias when fitting training videos.

\textbf{Action-aware Discrimination:}
Empirically, we find that although the Scene-aware Discrimination loss $L_{\text{scene}}$ effectively improves the cross-domain action recognition performance for the closed-set categories, it reduces the cross-domain performance for open-set categories. 
Thus, it leads to limited improvement in the overall open-vocabulary action recognition performance across domains.

A critical reason is that $L_{\text{scene}}$ pushes a video away from a scene-encoded text prompt in representation space, potentially resulting in some non-ground-truth action texts becoming more similar to the video than the scene-encoded prompt is. 
However, since the scene-encoded text prompt encodes semantic information of the ground-truth category (in addition to a specific scene), the video representation should be more dissimilar to non-ground-truth action texts than to the scene-encoded text prompt.
For example, consider a video that depicts a person playing basketball in a court. 
This video should have a more dissimilar representation to the text prompts of other categories (\eg, ``a video of a person kicking soccer.'') than to the scene-encoded text prompts (\eg, ``a video of a person playing basketball in the park.'').
Overall, this issue would cause some confusion in video representation space.

Accordingly, to alleviate the degradation of cross-domain open-set performance, we propose an Action-aware Discrimination loss to constrain video representation learning. 
Relative to the scene-encoded text representations that encodes ground-truth action semantics, our Action-aware Discrimination loss pushes the video representations away from the text representations of non-ground-truth categories, which is formulated as follows:
\begin{equation}
\begin{aligned}
    L_{\text{action}} = \sum_{n=1}^N \sum_{k\not=y}^{K} \frac{\max{\left(0, \delta - s(z_{x}, \tilde{z}^n_{a_y}) + s(z_{x}, z_{a_k}) \right)}}{N(K-1)},
    \label{eq:pull}
\end{aligned}
\end{equation}
where $\delta$ is a margin factor. 
In principle, $L_{\text{action}}$ introduces a constraint for non-ground-truth action text $a_k$ ($k\not=y$), \ie, $s(z_{x}, \tilde{z}^n_{a_y}) - s(z_{x}, z_{a_k}) \geq \delta$. 
By using both $L_{\text{scene}}$ and $L_{\text{action}}$, our proposed Scene-Aware Video-text Alignment learns a more reasonable video representation space, \ie, $s(z_{x}, z_{a_y}) > s(z_{x}, \tilde{z}^n_{a_y}) > s(z_{x}, z_{a_k})$ when $k\not=y$.

\subsection{Overall Training and Test}
The overall training loss of our XOV-Action model is given as follows:
\begin{equation}
\begin{aligned}
    L = L_{\text{vta}} + \lambda_{\text{eva}} L_{\text{eva}} + \lambda_{\text{scene}} L_{\text{scene}} + \lambda_{\text{action}} L_{\text{action}},
    \label{eq:total}
\end{aligned}
\end{equation}
where $\lambda_{\text{eva}}$, $\lambda_{\text{scene}}$ and $\lambda_{\text{action}}$ are non-negative coefficients for trade-off.

During test, for each video, we leverage both the global video representation $z_{x}$ and the top-$\hat{C}$ elaboration representations $\{\hat{z}_x^{i}\}$ for classification.
More specifically, we use an ensembled score for classification, which is formulated as follows:
\begin{equation}
\begin{aligned}
    p_k = \lambda_{\text{e}} s(z_{x}, z_{a_k}) +  \frac{1 - \lambda_{\text{e}}}{\hat{C}} \sum_{(i,j)\in O^k_{\hat{C}}} s(\hat{z}_x^{i}, f_{\text{txt}}(e^j_k)),
    \label{eq:score}
\end{aligned}
\end{equation}
where $p_k$ denotes the score of the $k$-th action category, and $\lambda_{\text{e}}\in[0, 1]$ is the trade-off coefficient for ensemble.
Finally, our model conducts the classification by identifying the highest value of $p_k$ across different action categories.

\section{The Proposed Cross-domain Open-Vocabulary Action Benchmark}
\label{sec:xovabench}
% \vskip -0.1in

Our work focuses on the generalizable open-vocabulary action recognition task, which aims to develop open-vocabulary action recognition models that are generalizable in \textit{unseen} target domains by training in the source domain, \ie, recognizing \textit{both closed-set and open-set action categories} in new test domains.
In order to evaluate models in this task, we establish XOVABench, the first CROSS-domain Open-Vocabulary Action recognition Benchmark, which provides a comprehensive way to analyze models across various situations. 
In what follows, we introduce the components of our XOVABench benchmark in detail, as well as evaluation metrics.

\subsection{Benchmark Components}
Our proposed XOVABench benchmark consists of two source datasets for training and four target datasets for test.
The two \textbf{source datasets} for training are as follows:

(1) \textit{Kinetics400}~\cite{DBLP:conf/cvpr/CarreiraZ17}: One of the most widely-used action recognition datasets, consisting of 400 action categories.
Videos in Kinetics400 are collected from YouTube, which are usually recorded in common daily environments (\eg, normal illumination and weather).
Existing open-vocabulary action recognition works typically use Kinetics400 for training, and we use the original split following them.

(2) \textit{Kinetics150}: A subset of Kinetics400, composed of 150 action categories selected from the full Kinetics400.
These 150 categories include all the closed-set categories of the four target datasets in comparison to Kinetics400, and the remaining categories are randomly sampled. We will illustrate the definition of closed-set categories later. 
We construct this subset to conduct detailed analysis for generalizable open-vocabulary action recognition models.

\begin{table}[t]
\fontsize{8.5pt}{14.5pt}\selectfont
\caption{Category statistics of four test domains in our XOVABench.
The closed-set and open-set categories are identified according to Kinetics400.
In the table, we also show the domain gap between each test domain and the training domain, as well as the quantitative measure of domain gaps. 
}
\vskip -0.05in
\label{tab:category}
\centering
% \hskip -0.1in
\resizebox{\linewidth}{!}{
\begin{tabular}{c | c  c  c  c}
\hline Test Domains         & UCF       & HMDB    & ARID      & NEC-Dr \\
\hline
\# of closed-set actions    & 50        & 33      & 6         & 7  \\
\# of open-set actions      & 51        & 18      & 5         & 9  \\
\# of all actions           & 101       & 51      & 11        & 16  \\
Domain gap                  & Small     & Moderate& Large     & Large  \\
Quantitative measure of gap & 0.429     & 0.626   & 0.789     & 0.850  \\
\hline
\end{tabular}
}
% \vskip -0.05in
\end{table}

The four \textbf{target datasets} for test are as follows: 

(1) \textit{UCF}~\cite{DBLP:journals/corr/abs-1212-0402}: One of the most widely-used action recognition datasets, consisting of 101 action categories.
Videos in UCF are collected from YouTube, and videos of each action are usually captured from specific or similar environments.
UCF has a \textit{small} domain gap compared with the Kinetics (source) domain, and previous open-vocabulary action recognition works~\cite{DBLP:conf/cvpr/RasheedK0KK23,DBLP:conf/icml/WengYLWJ23} commonly use UCF to evaluate their models' open-vocabulary recognition abilities.

(2) \textit{HMDB}~\cite{DBLP:conf/iccv/KuehneJGPS11}: A widely-used action recognition datasets consisting of 51 action categories.
Compared with UCF, videos in HMDB are captured from more unconstrained environments and more different camera views.
Specifically, Videos in HMDB are collected mainly from movies, and remaining videos are from Prelinger archive, YouTube or Google videos.
Overall, HMDB has a \textit{moderate} domain gap compared with the Kinetics domain~\cite{DBLP:conf/eccv/XuYCWWC22,DBLP:journals/tcsv/XuYCWWLC23}.

(3) \textit{ARID}~\cite{xu2020arid}: A dataset consisting of 11 categories of action videos, which are recorded under dark environments.
These actions include singular person actions (\eg, jump, run) and actions associated with objects (\eg, drink, pick).
Due to the significantly different illumination conditions, ARID exhibits a \textit{large} domain gap compared with the Kinetics domain.
When training data are limited to videos with normal illumination, developing models that are generalizable in ARID is challenging.

(4) \textit{NEC-Dr}~\cite{DBLP:conf/wacv/ChoiSCH20}: A dataset consisting of 16 categories of action videos, which are recorded by drones in the same basketball court.
These actions include single-person actions (\eg, jump, walk) and interactive actions (\eg, hug, shake hands).
Due to the different shooting equipments and scenarios, NEC-Dr exhibits a \textit{large} domain gap compared with the Kinetics domain.
Therefore, it is challenging to make a Kinetics-trained model generalizable in such a drone video domain.

These test domains exhibit different levels of domain gaps in comparison to the training domain. 
To quantitatively measure the domain gaps, we use the mean of class-wise feature discrepancy of closed-set categories as the evaluation metric following previous works~\cite{DBLP:conf/icml/LongC0J15,DBLP:conf/eccv/SunS16}. 
A summary of four test domains is given in Table~\ref{tab:category}.
In our experiments, we use videos from Kinetics400 or Kinetics150 for training.
Then, we conduct evaluation on the four target datasets to assess open-vocabulary action recognition abilities across domains.

\subsection{Evaluation Metrics}
First, we illustrate the definitions of \textit{closed-set} and \textit{open-set} categories for test domains:
(a) The closed-set categories refer to the categories that share similar meanings and have a common lexicon with the categories in the training domain, following a similar approach in previous zero-shot action recognition works~\cite{DBLP:conf/cvpr/BrattoliTZPC20,DBLP:conf/iccv/ChenH21}. 
(b) The open-set categories refer to the remaining categories that are not involved in the training domain.
Then, we identify closed-set and open-set categories for each test domain, according to the Kinetics400 categories.
Category statistics are summarized in {Table~\ref{tab:category}}. 

In our experiments, we adopt three evaluation metrics:
(1) The \textit{closed-set accuracy} measures the recognition performance of closed-set categories, which primarily evaluates the model abilities of bridging domain gaps when fitting training videos.
(2) The \textit{open-set accuracy} measures the performance of open-set categories, which evaluates the generalization abilities across both video domains and action categories.
(3) The \textit{overall accuracy} measures the recognition performance over all categories, which provides a holistic view of model effectiveness across various situations.

\begin{table*}[t]
% \vskip -0.1in
\fontsize{8.5pt}{14.5pt}\selectfont
\setlength\tabcolsep{4.0pt}
\caption{
Comparison with Kinetics150-trained open-vocabulary action recognition models on XOVABench. 
``AVG'' denotes the average ACC over four test domains.
The bold/underlined numbers indicate the best/second best.
We prioritize the average overall ACC as the primary metric in model comparison.
To conduct solid comparisons, we carefully tune compared models based on their official code and select the checkpoint with the best average overall ACC for each model for comparison. 
}
\label{tab:k150}
% \vskip -0.05in
\centering
% \hskip -0.1in
\begin{tabular}{c || c  c  c | c c c | c c c | c c c | c c c }
% \toprule[0.5pt]
\hline
\multirow{2}{*}{Models}
& \multicolumn{3}{c|}{\textbf{UCF}}
& \multicolumn{3}{c|}{\textbf{HMDB}}
& \multicolumn{3}{c|}{\textbf{ARID}}
& \multicolumn{3}{c|}{\textbf{NEC-Dr}}
& \multicolumn{3}{c}{\textbf{AVG}}   \\
\cline{2-16}
& Closed    & Open      & All      & Closed    & Open      & All    & Closed    & Open      & All   & Closed    & Open      & All   & Closed    & Open      & All      \\
% \midrule[0.5pt]
\hline
CLIP~\cite{DBLP:conf/icml/RadfordKHRGASAM21}
& 61.51     & 57.85  & 59.67  & 42.92     & 38.62   & 41.40 & 32.04    & 12.04  & 24.02 & 18.52     & 6.44    & 12.33  & 38.75     & 28.74  & 34.35      \\
~~~ActionCLIP~\cite{DBLP:journals/corr/abs-2109-08472}~~~
& 84.76     & 56.27  & 70.44  & 56.78     & 33.02   & 48.40 & 42.25    & 19.75  & 33.23 & 18.75     & 4.22    & 11.30  & 50.63     & 28.32     & 40.84   \\
LSTM~\cite{DBLP:journals/corr/abs-2109-08472}
& 84.54     & 58.38  & 71.39  & 58.30     & 36.38   & 50.56 & 42.04    & 26.85  & 35.95 & 17.13     & 12.44   & 14.73  & 50.50     & 33.51     & 43.16   \\
TConv~\cite{DBLP:journals/corr/abs-2109-08472}
& 84.54     & 58.61  & 71.24  & 57.29     & 36.19   & 49.84 & 44.29    & 27.78  & 37.67 & 17.59     & 12.22   & 14.84  & 50.93     & 33.56     & 43.40   \\
XCLIP~\cite{DBLP:conf/eccv/NiPCZMFXL22}
& \textbf{89.89}     & 59.38  & 74.59  & 53.44     & 30.78   & 45.62 & 35.10    & 13.72  & 26.64 & 16.90     & 5.70    & 11.26  & 48.83     & 27.40     & 39.53   \\
Text4Vis~\cite{DBLP:conf/aaai/WuSO23}
& 83.40     & 49.84  & 66.54  & 54.25     & 38.06   & 48.54 & 41.43    & \underline{31.71}  & 37.53 & 12.50     & 4.17    & 8.23   & 47.90     & 30.94     & 40.21   \\
{BIKE}~\cite{DBLP:conf/cvpr/WuWLWYO23} 
& 74.36     & 55.80  & 65.03  & 50.71     & 38.79   & 46.50 & 36.09    & 16.67  & 28.30 & 17.03     & 13.16   & 15.04  & 44.55     & 31.10     & 38.72   \\
ViFiCLIP~\cite{DBLP:conf/cvpr/RasheedK0KK23} 
& 82.93     & 62.07  & 72.44  & 58.60     & 41.60   & \underline{52.60} & 44.69    & 27.13  & 37.65 & 17.36     & 10.53   & 13.86  & 50.90     & 35.33     & 44.14   \\
{OpenVCLIP}~\cite{DBLP:conf/icml/WengYLWJ23} 
& 87.67     & \underline{70.86}  & 79.20  & 56.48     & \underline{44.96}   & 52.41 & \underline{48.16}    & 27.13  & \underline{39.73} & \underline{19.68}     & 13.38   & \underline{16.45}  & \underline{53.00}     & \underline{39.08}     & \underline{46.95}   \\
{FROSTER}~\cite{DBLP:journals/corr/abs-2402-03241} 
& \underline{89.80}     & 69.17  & \underline{79.43}  & 53.94     & 35.19   & 47.32 & 39.15    & 29.57  & 35.31 & 13.13     & 1.97   & 7.41  & 49.01     & 33.98     & 42.37   \\
{TC-CLIP}~\cite{DBLP:conf/eccv/KimHKH24} 
& 87.41     & 31.67  & 59.40  & 49.39     & 32.41   & 43.40 & 47.06    & 13.33  & 33.54 & 19.59     & 10.07   & 14.70  & 50.86     & 21.89     & 37.76   \\
{MoTE}~\cite{DBLP:nips/abs-2410-10589} 
& 85.19     & 61.51  & 73.29  & \underline{58.69}     & 32.96   & 49.61 & 44.85    & 16.36  & 33.43 & 14.48     & \underline{15.49}   & 14.99  & 50.80     & 31.58     & 42.83   \\
\hline
\rowcolor{pink!20}
XOV-Action (Ours)
& 88.46     & \textbf{72.02}  & \textbf{80.20}  & \textbf{60.22}     & \textbf{46.46}   & \textbf{55.36} & \textbf{50.41}    & \textbf{32.01}  & \textbf{43.03} & \textbf{24.77}     & \textbf{16.67}   & \textbf{20.61}  & \textbf{55.96}     & \textbf{41.79}     & \textbf{49.80} \\
\hline
\end{tabular}
\end{table*}

Note that, previous open-vocabulary action recognition works~\cite{DBLP:conf/cvpr/RasheedK0KK23,DBLP:conf/icml/WengYLWJ23} usually treat all the categories in UCF and HMDB as open-set for models trained on Kinetics400. 
However, UCF and HMDB share many overlapping categories with Kinetics400. For example, UCF has 50 categories that overlap with Kinetics400. Thus, these works make an inaccurate assessment of models' open-vocabulary abilities. 
Unlike these works, we provide a more accurate way for evaluation by identifying closed-set and open-set categories according to the training dataset.
By evaluating across various domains and different action categories, our proposed XOVABench benchmark enables us to perform a wide range of analysis for generalizable open-vocabulary action recognition models.
To achieve satisfactory performance on our proposed XOVABench, models should consider the challenges of both novel action concepts and domain gaps, since XOVABench consists of test domains containing various action categories and exhibiting different levels of domain gaps.

\section{Experiment Results}

\subsection{Experimental Setups}

\textbf{Implementation Details:}
For each video, our model takes $T=16$ frames of size $224\times 224$ as inputs.
Following ViFiCLIP~\cite{DBLP:conf/cvpr/RasheedK0KK23}, we adopt temporal jitter, multi-scale random spatial crop and color jitter for augmentation during training.
During test, we use one temporal clip composed of center-cropped frames for inference.
Regarding the architecture of the video encoder, we adopt ViT-B/32 and ViT-B/16 initialized by CLIP image encoder for Kinetics150 and Kinetics400, respectively.
For temporal modeling, we set the temporal receptive field of self-attention layers to 7 in the video encoder. 
By default, we learn {$C=10$} video elaboration modules in training, and select {$\hat{C}=3$} elaboration representations for each video for both training and inference. 
These video elaboration modules are instantiated using MLPs with residuals, and thus they are lightweight and have a negligible impact on computational efficiency. 
We ask GPT-4~\cite{openai2023gpt4} to output 300 scene suffixes that are common in daily life, and randomly sample $N=50$ scene suffixes for our proposed Scene-Aware Video-text Alignment in each batch.
In addition, we ask GPT-4 to output {$M=10$} textual descriptions for each action category in each dataset.
We freeze the text encoder during training and thus all text epresentations can be pre-computed before training.
During inference, we set {$\lambda_{\text{e}}=0.4$} to obtain an ensembled scores for action classification. 
During training, network parameters are optimized using AdamW optimizer with a batch size of 512, a learning rate of 8e-6, a momentum of 0.9 and a weight decay of 1e-3.
If not specified, we set the loss coefficients as {$\lambda_{\text{scene}}=0.2$, $\lambda_{\text{action}}=0.1$, $\lambda_{\text{eva}}=0.1$, the margin as $\delta=0.5$,} and the temperature coefficient as $\tau=0.01$.
Following OpenVCLIP~\cite{DBLP:conf/icml/WengYLWJ23}, we adopt the SWA technique~\cite{DBLP:conf/uai/IzmailovPGVW18} for producing the final model.

\textbf{Evaluation Protocol:}
We adopt nine state-of-the-art open-vocabulary action recognition models, which have released their codes, to conduct model comparisons.
To conduct solid comparisons, we carefully tune each model on Kinetics150/Kinetics400 based on the official code, and report the best generalizable open-vocabulary action recognition performance for each model. 
For a fair comparison, we adopt consistent backbones and same training recipes for all models (\eg, epoch number), and we use one temporal clip during inference.

We evaluate each model on all the four test domains of our XOVABench benchmark, where different test domains have various levels of domain gaps and differ in action category. 
For a comprehensive analysis, we report the accuracy of closed-set, open-set and all categories in individual test domains as well as the average accuracies.
When conducting a comparison between different models, we prioritize the average overall accuracy as the primary metric.

\begin{table*}[t]
% \vskip -0.1in
\fontsize{8.5pt}{14.5pt}\selectfont
\setlength\tabcolsep{4.0pt}
\caption{
Comparison with Kinetics400-trained open-vocabulary action recognition models on XOVABench. 
``AVG'' denotes the average ACC over four test domains.
The bold/underlined numbers indicate the best/second best.
We prioritize the average overall ACC as the primary metric in model comparison. 
To conduct solid comparisons, we carefully tune compared models based on their official code and select the checkpoint with the best average overall ACC for each model for comparison.
}
\label{tab:k400}
% \vskip -0.05in
\centering
\begin{tabular}{c || c  c  c | c c c | c c c | c c c | c c c }
\hline
\multirow{2}{*}{Models}
& \multicolumn{3}{c|}{\textbf{UCF}}
& \multicolumn{3}{c|}{\textbf{HMDB}}
& \multicolumn{3}{c|}{\textbf{ARID}}
& \multicolumn{3}{c|}{\textbf{NEC-Dr}}
& \multicolumn{3}{c}{\textbf{AVG}}   \\
\cline{2-16}
& Closed    & Open      & All      & Closed    & Open      & All    & Closed    & Open      & All   & Closed    & Open      & All   & Closed    & Open      & All      \\
\hline
CLIP~\cite{DBLP:conf/icml/RadfordKHRGASAM21}
& 66.71     & 59.53  & 63.15  & 49.19     & 35.26   & 44.27 & 37.76    & 12.65  & 27.69 & 10.42     & 22.44   & 16.59  & 41.04     & 32.47     & 37.92   \\
{ActionCLIP}~\cite{DBLP:journals/corr/abs-2109-08472}~~~
& 89.87     & 51.42  & 70.55  & 64.07     & 23.32   & 49.69 & 48.98    & 22.53  & 38.38 & 19.91     & 8.89     & 14.26  & 55.71     & 26.54     & 43.22   \\
{XCLIP}~\cite{DBLP:conf/eccv/NiPCZMFXL22}
& 89.50     & 60.96  & 75.16  & 60.32     & 36.01   & 51.74 & 45.51    & 18.21  & 34.56 & 22.22     &  9.78    & 15.84  & 54.39     & 31.24     & 44.33   \\
{Text4Vis}~\cite{DBLP:conf/aaai/WuSO23}
& 87.21     & 54.16  & 70.60  & 59.31     & 26.31   & 47.66 & 42.44    & 21.30  & 33.97 & 15.97     & 14.67    & 15.30  & 51.24     & 29.11     & 41.88   \\
{BIKE}~\cite{DBLP:conf/cvpr/WuWLWYO23} 
& 81.04     & 61.12  & 71.03  & 53.63     & 39.52   & 48.65 & 46.98    & 15.18  & 34.23 & 23.92     & 10.94   & 17.26   & 51.39     & 31.69     & 42.79   \\
{ViFiCLIP}~\cite{DBLP:conf/cvpr/RasheedK0KK23} 
& 90.51     & 56.27  & 73.31  & \underline{64.98}     & 30.22   & 52.71 & 51.92    & 19.14  & 38.77 & 21.99     & 11.78   & 16.75  & \underline{57.35}     & 29.35     & 45.39   \\
{OpenVCLIP}~\cite{DBLP:conf/icml/WengYLWJ23} 
& \underline{91.06}     & \underline{75.03}  & \underline{83.00}  & 60.63     & \underline{42.54}   & 54.24 & 52.25    & 28.35  & \underline{42.67} & 22.92     & 14.91   & \underline{18.81}   & 56.71     & 40.21     & \underline{49.68} \\
{FROSTER}~\cite{DBLP:journals/corr/abs-2402-03241} 
& 89.53     & 74.86  & 82.16  & 59.70     & 38.89   & 52.36 & 47.06    & \textbf{33.63}  & 41.67 & 11.06     & \underline{16.19}   & 13.69   & 51.84     & \underline{40.89}      & 47.47   \\
{TC-CLIP}~\cite{DBLP:conf/eccv/KimHKH24} 
& 90.91     & 67.23  & 79.01  & 60.20     & 32.22   & 50.33 & \underline{52.74}    & 23.64  & 41.07 & \underline{24.19}     & 8.75   & 16.27  & 57.01     & 32.96     & 46.67   \\
{MoTE}~\cite{DBLP:nips/abs-2410-10589} 
& 86.78     & 67.04  & 76.86  & 61.62     & 41.67   & \underline{54.58} & 50.51    & 22.73  & 39.37 & 12.87     & 15.32   & 14.13  & 52.94     & 36.69     & 46.23   \\
\hline
\rowcolor{pink!20}
XOV-Action (Ours)
& \textbf{91.97}     & \textbf{80.08}  & \textbf{86.00}  & \textbf{65.28}     & \textbf{43.83}   & \textbf{57.72} & \textbf{54.29}    & \underline{32.32}  & \textbf{45.48} & \textbf{24.31}     & \textbf{16.67}   & \textbf{20.39}  & \textbf{58.96}     & \textbf{43.23}     & \textbf{52.39}\\
\hline
\end{tabular}
\vskip -0.0in
\end{table*}

\subsection{Comparison with State-of-the-arts}
The results are summarized in Table~\ref{tab:k150} and \ref{tab:k400}.
As shown in the tables, by adapting CLIP to video data, previous open-vocabulary action recognition models obtain substantial improvement over the original CLIP for generalizable open-vocabulary action recognition. 
Although these methods show impressive performance for the domain with a small gap (UCF), they still exhibit limited performance for the domains with large gaps (ARID and NEC-Dr). 
These results reveal potential challenges of the task, as discussed in Section~\ref{sec:intro}.  
By learning scene-agnostic video representations and diversified elaboration representations, our proposed XOV-Action outperforms previous state-of-the-arts by {2.85\% and 2.71\%} in terms of average overall accuracy for Kinetics150 and Kinetics400, respectively. 
These improvements are significant, especially considering that the metric is an average across four test domains.
More importantly, our proposed XOV-Action obtains the best performance in most cases on XOVABench, demonstrating its superior capabilities across different category types and different test domains. 
This success is attributed to our effective designs specified in open-set action understanding and cross-domain generalization.

\begin{table}[t]
% \vskip -0.05in
\fontsize{8.5pt}{14.5pt}\selectfont
\setlength\tabcolsep{4.0pt}
\caption{
Comparison with state-of-the-art open-vocabulary action recognition models on Kinetics600 zero-shot (K600-ZS) and Kinetics400 base-to-novel (K400-B2N) benchmarks. 
The bold/underlined numbers indicate the best/second best.
}
\label{tab:generalization}
\vskip -0.05in
\centering
\begin{tabular}{c || c | c c c  }
% \toprule[0.5pt]
\hline
\multirow{2}{*}{Models}
& \multicolumn{1}{c|}{\textbf{~~~~K600-ZS~~~~}}
& \multicolumn{3}{c}{\textbf{K400-B2N}}   \\
\cline{2-5}
& Top-1     & ~~~Base    & ~~Novel~~   & HM~~~   \\
% \midrule[0.5pt]
\hline
CLIP~\cite{DBLP:conf/icml/RadfordKHRGASAM21}
& 68.1$_{\pm 1.1}$  & ~~~62.3  & 53.4  & 57.5~~~  \\
% \hline
ActionCLIP~\cite{DBLP:journals/corr/abs-2109-08472}
& 62.5$_{\pm 1.2}$  & ~~~61.0  & 46.2  & 52.6~~~ \\
VPT~\cite{DBLP:conf/eccv/JuHZZX22}
& 55.8$_{\pm 0.7}$  & ~~~69.7  & 37.6  & 48.8~~~ \\
XCLIP~\cite{DBLP:conf/eccv/NiPCZMFXL22}
& 65.2$_{\pm 0.4}$  & ~~~74.1  & 56.4  & 64.0~~~ \\
AIM~\cite{DBLP:conf/iclr/YangZXZC023}
& 66.7$_{\pm 0.5}$  & ~~~74.6  & 62.5  & 68.0~~~ \\
ST-Adapter~\cite{DBLP:conf/nips/PanLZS022}
& 60.2$_{\pm 1.8}$  & ~~~73.6  & 62.0  & 67.3~~~ \\
ViFiCLIP~\cite{DBLP:conf/cvpr/RasheedK0KK23}
& 71.2$_{\pm 1.0}$  & ~~~76.4  & 61.1  & 67.9~~~ \\
OpenVCLIP~\cite{DBLP:conf/icml/WengYLWJ23}
& 73.0$_{\pm 0.8}$  & ~~~76.5  & 62.6  & 68.9~~~ \\
FROSTER~\cite{DBLP:journals/corr/abs-2402-03241}
& \underline{74.8}$_{\pm 0.9}$  & ~~~\underline{77.8}  & \underline{64.3}  & \underline{70.4}~~~ \\
\rowcolor{pink!20}
~~~XOV-Action (Ours)~~~
& \textbf{76.5}$_{\pm 0.8}$  & ~~~\textbf{79.0}    & \textbf{66.0}    & \textbf{71.9}~~~ \\
% \bottomrule[0.5pt]
\hline
\end{tabular}
\vskip -0.1in
\end{table}

In addition to XOVABench, we conduct experiments on Kinetics600 zero-shot benchmark and Kinetics400 base-to-novel benchmark to demonstrate the generalization capabilities of our proposed XOV-Action model. 
Specifically, for the Kinetics600 zero-shot benchmark, models are trained on Kinetics400 and evaluated on Kinetics600's extra action categories, \ie, novel categories that are not included in Kinetics400. 
For the Kinetics400 base-to-novel benchmark, models are trained on the base categories of Kinetics400 and evaluated on the remaining (novel) categories.
Following the standard protocol~\cite{DBLP:conf/cvpr/RasheedK0KK23,DBLP:conf/icml/WengYLWJ23}, we report the Top-1 accuracy of models on the Kinetics600 zero-shot benchmark, and report the accuracies for both base and novel categories and the Harmonic Mean (HM) for an overall measurement. 
As shown in Table~\ref{tab:generalization}, our XOV-Action model significantly outperforms previous state-of-the-art models by {1.7\% and 1.5\%} on these two benchmarks, respectively. 
These results demonstrate the strong generalization capabilities of our model.

\begin{table*}[t]
\vskip -0.0in
\fontsize{8.5pt}{14.5pt}\selectfont
\setlength\tabcolsep{4.0pt}
\caption{
Ablation study of our proposed XOV-Action trained on Kinetics150. 
``AVG'' denotes the average ACC over four test domains.
The closed-set accuracy primarily evaluates the model abilities of tackling domain gaps when fitting training videos.
}
\label{tab:ablation}
\vskip -0.05in
\centering
\begin{tabular}{c || c  c  c | c c c | c c c | c c c | c c c }
% \toprule[0.5pt]
\hline
\multirow{2}{*}{Models}
& \multicolumn{3}{c|}{\textbf{UCF}}
& \multicolumn{3}{c|}{\textbf{HMDB}}
& \multicolumn{3}{c|}{\textbf{ARID}}
& \multicolumn{3}{c|}{\textbf{NEC-Dr}}
& \multicolumn{3}{c}{\textbf{AVG}}   \\
\cline{2-16}
% & C         & O      & A      & C         & O       & A     & C        & O      & A     & C         & O       & A      & C         & O      & A     \\
& Closed    & Open      & All      & Closed    & Open      & All    & Closed    & Open      & All   & Closed    & Open      & All   & Closed    & Open      & All      \\
% \midrule[0.5pt]
\hline
$L_{\text{vta}}$
& 87.07     & 69.39  & 78.19  & 57.59     & 44.59   & 53.00 & 47.35    & 26.52  & 39.00 & 21.30     & 14.04   & 17.57  & 53.33     & 38.64     & 46.94   \\
+$L_{\text{scene}}$
& 87.82     & 69.07  & 78.40  & 59.41     & 43.10   & 53.65 & 49.18    & 25.31  & 39.61 & 25.93     & 10.53   & 18.03  & 55.59     & 37.00     & 47.42 \\
+$L_{\text{scene}}$\&$L_{\text{action}}$
& 87.98     & 69.97  & 78.93  & 58.40     & 44.59   & 53.53 & 48.78    & 29.57  & 41.08 & 24.54     & 12.72   & 18.48  & 54.92     & 39.21     & 48.00 \\
\rowcolor{pink!20}
Full
& 88.46     & 72.02  & 80.20  & 60.22     & 46.46   & 55.36 & 50.41    & 32.01  & 43.03 & 24.77     & 16.67   & 20.61  & 55.96     & 41.79     & 49.80 \\
% \bottomrule[0.5pt]
\hline
\end{tabular}
\end{table*}

\subsection{Main Ablation Study}
Table~\ref{tab:ablation} summarizes the results of our ablation study.
In this experiment, we use the model trained with only the video-text alignment loss $L_{\text{vta}}$ as the baseline.
By introducing our proposed Scene-aware Discrimination loss $L_{\text{scene}}$, our model obtains improvement on XOVABench in terms of closed-set accuracy on all the four test domains.
Specifically, our model obtains significant improvements of {1.83\% and 4.63\%} in terms of closed-set accuracy on ARID and NEC-Dr, which have large domain gaps with the training domain.
This is because that $L_{\text{scene}}$ encourages the video encoder to downweight the attention on scene information in videos and thus pay more attention to action information that are generalizable across different domains.
However, we find that introducing only the Scene-aware Discrimination loss reduces the open-set performance, leading to limited improvement in average overall accuracy on XOVABench.
By introducing the Action-aware Discrimination loss $L_{\text{action}}$, our model can obtain improved cross-domain closed-set accuracy over the baseline while satisfactorily maintaining the open-set accuracy on each test domain, and this leads to an improvement of {1.06\%} in terms of average overall accuracy over the baseline.
Furthermore, by introducing our proposed Diversified Elaboration Representation Learning, our full XOV-Action model can obtain significant improvement for the open-set action categories on each test domain, leading to a large improvement of {2.58\%} on avarage. 
This is attributed to the capture of rich action-related concepts in videos. 
Overall, our model can effectively improve the generalization of both closed-set and open-set action categories in unseen test domains.

\subsection{Quantitative Analysis of Diversified Elaboration Representation Learning}
\subsubsection{Quantitative Analysis of the Module and Loss Designs}
In this part, we conduct a quantitative analysis to the module and loss design of our proposed Diversified Elaboration Representation Learning. 
We use our XOV-Action without Diversified Elaboration Representation Learning as the baseline, which refers to the model using only Scene-Aware Video-text Alignment. 
As shown in Table~\ref{tab:diverse_elaboration}, by introducing extra textual descriptions in inference, the baseline obtains better open-set action recognition performance compared with the one using only category names, \ie, from {39.21\% to 40.56\%}. 
This is because that, textual descriptions encodes much more concept information about the corresponding action category than simple name texts, and thus model can better associate videos with open-set categories during testing. 
However, introducing textual descriptions in only inference obtains limited performance improvement in open-set generalization. 
Therefore, we propose to introduce textual descriptions and conduct alignment between videos and textual descriptions during training. 
As shown by the ``Full'' in the table, our proposed Diversified Elaboration Representation Learning can effectively improve the open-set action recognition performance, and {also modestly improve} the closed-set recognition. 
This demonstrates that learning diverse action concepts can effectively boost the understanding of novel action concepts of open-set categories, thus improving the open-set action recognition performance. 

\begin{table}[t]
\vskip -0.0in
%\small
\fontsize{8.5pt}{14.5pt}\selectfont
\caption{
Quantitative analysis of the module and loss designs of our proposed Diversified Elaboration Representation Learning. 
In this table, we report the average closed-set and open-set performance for each model on XOVABench. 
For all models, we used both category names and textual descriptions for action classification during inference. 
}
\label{tab:diverse_elaboration}
\vskip -0.05in
\centering
\begin{tabular}{c | c | c c }
\hline Loss        & Model      & ~Closed-set~    & ~Open-set~   \\
\hline
~w/o $L_{\text{eva}}$~           & -         & 55.06           & 40.56      \\
\multirow{3}{*}{$L$ (Eq.~\eqref{eq:total})} 
& w/o elaboration modules        & 55.78           & 40.85    \\
& w/o adaptive matching          & 54.97           & 41.13    \\
& Full                           & 55.96           & 41.79    \\
\hline
\end{tabular}
\vskip -0.1in
\end{table}

In addition, results in Table~\ref{tab:diverse_elaboration} demonstrate the importance of Adaptive Elaboration Matching formulated by Eq.~\eqref{eq:matching}. 
Specifically, if we remove Adaptive Elaboration Matching from our model and follow the index order of descriptions to supervise elaboration representation learning, both closed-set and open-set performance decreases, as shown by the ``w/o adaptive matching''. 
This is because the video elaboration modules in our model are not order-sensitive, so following the index order of descriptions to learn elaboration representations will result in confused representation space. 
In addition, we also analyze the case where video elaboration modules are removed, \ie, the loss $L_{\text{eva}}$ is directly applied on the global video representation $f_{\text{vid}}(x)$ without extra specialized module designs. 
In this case, open-set performance decreases, which demonstrates the effectiveness of video elaboration modules for learning diverse action concepts.

\subsubsection{Effect of Elaborative Video-text Alignment}
\begin{figure}[t]
\begin{minipage}[t]{0.46\textwidth}
\centering
\includegraphics[width=0.98\linewidth]{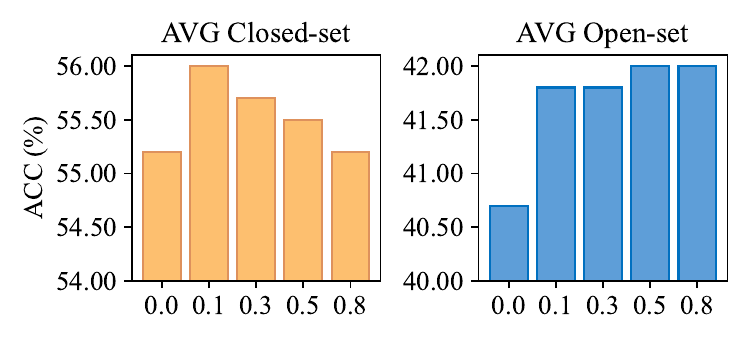}
\vskip -0.15in
\caption{
Quantitative analysis of the coefficient $\lambda_{\text{eva}}$ of our Elaborative Video-text Alignment loss.
By default, we set $\lambda_{\text{eva}}=0.1$ by default according to the best trade-off between closed-set and open-set recognition. 
}
\label{fig:exp_elaborative_loss_weight}
\end{minipage}
\vskip -0.15in
\end{figure}

We conduct a quantitative analysis to the loss coefficient $\lambda_{\text{eva}}$ in our proposed Elaborative Video-text Alignment loss, and the results are shown in Figure~\ref{fig:exp_elaborative_loss_weight}.
As shown in the figure, by introducing our Elaborative Video-text Alignment loss, the open-set action recognition performance gets significant improvement, \ie, coefficient 0.0 vs. 0.1. 
These results demonstrate the effectiveness of learning diverse action concepts. 
As the loss coefficient goes larger, the open-set performance gets gradually higher, as the model focuses more on learning extra action concepts encoded by the textual descriptions. 
However, a large loss coefficient $\lambda_{\text{eva}}$ will hinder the fitting of closed-set action categories during training, thus the closed-set performance gets gradually lower as the $\lambda_{\text{eva}}$ goes larger.  
Overall, we set $\lambda_{\text{eva}}=0.1$ by default according to the best trade-off between closed-set and open-set recognition. 

\begin{figure}[t]
\centering
\includegraphics[width=0.95\linewidth]{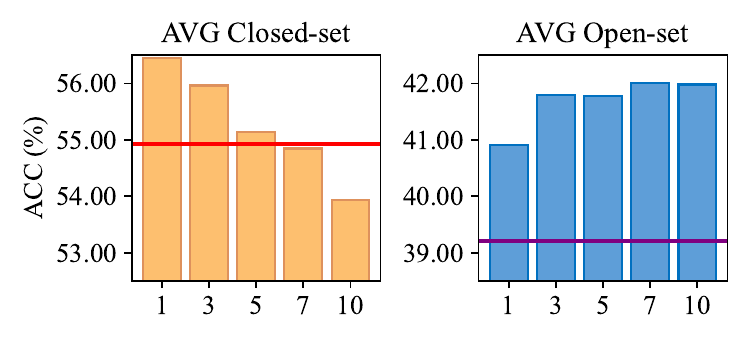}
\vskip -0.15in
\caption{
Quantitative analysis of the number of selected elaboration representations $\hat{C}$ for training and inference.
In this experiment, we use our XOV-Action without Diversified Elaboration Representation Learning as the baseline. 
In the two subfigures, the red and purple lines denotes the average closed-set and open-set performance of the baseline, respectively. 
}
\label{fig:exp_selected_elaboration}
\vskip -0.1in
\end{figure}

\begin{figure}[t]
\begin{minipage}[t]{0.23\textwidth}
\centering
\includegraphics[width=1\linewidth]{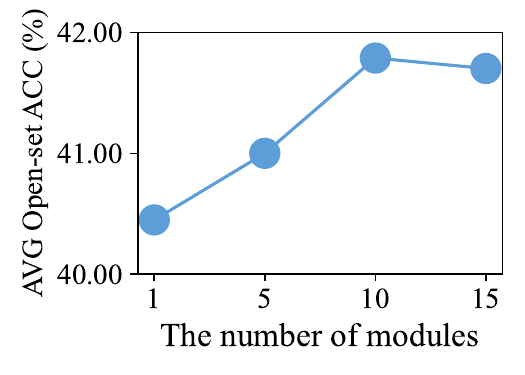}
\vskip -0.1in
\caption{
Quantitative analysis of the number of video elaboration modules.
Our model performs generally better with more video elaboration modules. 
}
\label{fig:exp_number_elaboration}
\end{minipage}
\hskip 0.1in
\begin{minipage}[t]{0.23\textwidth}
\centering
\includegraphics[width=1\linewidth]{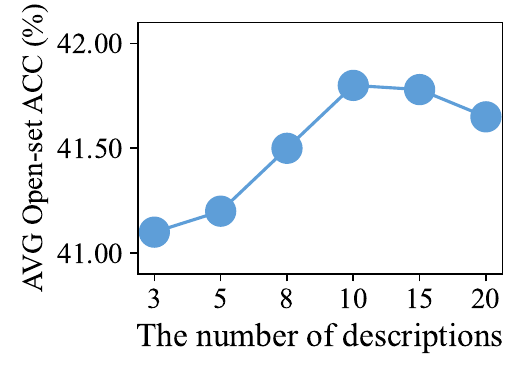}
\vskip -0.1in
\caption{
Quantitative analysis of the number of textual descriptions for each action category.
Our model performs generally better with more textual descriptions. 
}
\label{fig:exp_num_desc}
\end{minipage}
\end{figure}

\subsubsection{Effect of Confidence-aware Elaboration Selection}
In Diversified Elaboration Representation Learning, we propose a confidence-aware elaboration selection strategy to select the top-$\hat{C}$ most relevant descriptions for each category, which aims to mitigate the negative effects of noisy textual descriptions during training. 
In this part, we conduct a quantitative analysis to the number of selected elaboration representations $\hat{C}$ in our confidence-aware elaboration selection, and the results are shown in Figure~\ref{fig:exp_selected_elaboration}.
As shown in the figure, our proposed Diversified Elaboration Representation Learning consistently improves open-set action recognition using different number of selected elaboration representations, compared with our model without Diversified Elaboration Representation Learning (\ie, the purple baseline in the figure). 
In addition, the results show that open-set performance is not sensitive to the number of selected elaboration representations when $\hat{C}\geq 3$. 
However, selecting a large number of elaboration representations leads to a noticeable performance drop in closed-set recognition. 
This is because GPT-generated textual descriptions may include incorrect or video-irrelevant details, and using a large number of elaboration representations introduces noise when fitting closed-set action videos. 
By default, we set $\hat{C}=3$, which provides the best trade-off between closed-set and open-set action recognition.

\subsubsection{Effect of Video Elaboration Modules}
In Figure~\ref{fig:exp_number_elaboration}, we show how the number of video elaboration modules affects the model performance. 
As shown in the figure, our model generally obtains better open-set accuracy with more video elaboration modules introduced. 
This is attributed to that our proposed Diversified Elaboration Representation Learning can better capture diverse action-related concepts when using more video elaboration modules, according to the guidance of textual descriptions. 
Empirically, we find that matching the number of video elaboration modules to the number of textual descriptions is an appropriate hyperparameter choice (\ie, $C=M=10$), and our results show that using too many video elaboration modules may not yield performance improvement. 

\begin{figure}[t]
\vskip -0.1in
\centering
\includegraphics[width=0.95\linewidth]{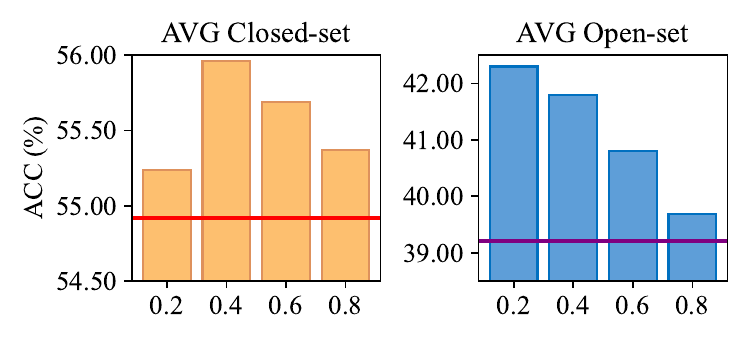}
\vskip -0.15in
\caption{
Quantitative analysis of the ensemble coefficient $\lambda_{\text{e}}$ for action classification. 
In this experiment, we set the the baseline as our XOV-Action without Diversified Elaboration Representation Learning. 
In the two subfigures, the red and purple lines denotes the average closed-set and open-set performance of the baseline, respectively. 
}
\label{fig:exp_ensemble_weight}
\vskip -0.1in
\end{figure}

\subsubsection{Effect of the Number of Textual Descriptions}
In Figure~\ref{fig:exp_num_desc}, we show how the number of textual descriptions affects the model performance. 
As shown in the figure, our model obtains better open-set accuracy when more textual descriptions introduced. 
This is intuitive since more textual descriptions contain more action-related details, and thus our proposed Diversified Elaboration Representation Learning can learn rich action concepts. 
As shown in the figure, when using a very large number of textual descriptions, the open-set performance decreases slightly. 
This is because the textual descriptions are automatically generated by GPT, and it will introduce noise more than action-related concept information when too many textual descriptions are generated.

\begin{figure*}[t]
\begin{minipage}[t]{0.7\textwidth}
\centering
\includegraphics[width=1.0\linewidth]{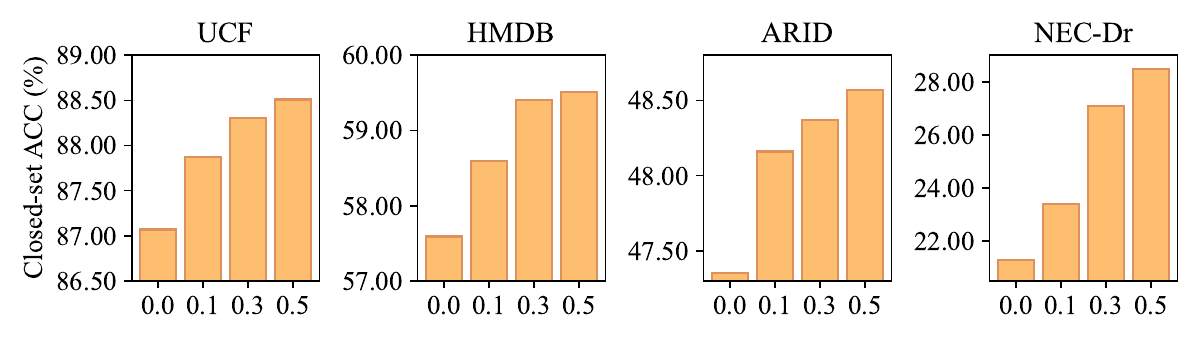}
\vskip -0.15in
\caption{Quantitative analysis of the coefficient $\lambda_{\text{scene}}$ for the Scene-aware Discrimination loss by closed-set accuracy on four test domains.
The horizontal axis shows the value of $\lambda_{\text{scene}}$.
Our model consistently obtains better closed-set performance on all test domains with a larger loss weight. 
}
\label{fig:exp_dis}
\end{minipage}
\hskip 0.2in
\begin{minipage}[t]{0.28\textwidth}
\centering
\includegraphics[width=0.98\linewidth]{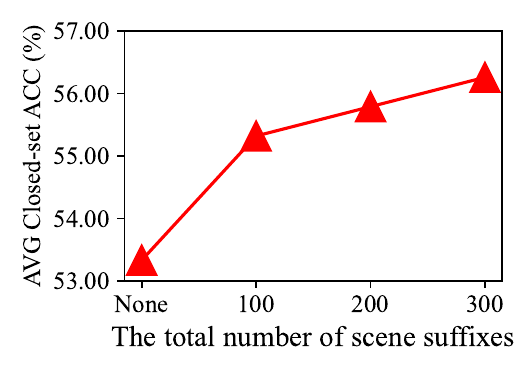}
\vskip -0.15in
\caption{
Quantitative analysis of the total number of scene suffixes.
Our model obtains better closed-set accuracy with more scene suffixes. 
}
\label{fig:exp_scene}
\end{minipage}
\end{figure*}

\begin{figure*}[t]
\vskip -0.1in
\begin{minipage}[t]{0.49\textwidth}
\centering
\includegraphics[width=0.95\linewidth]{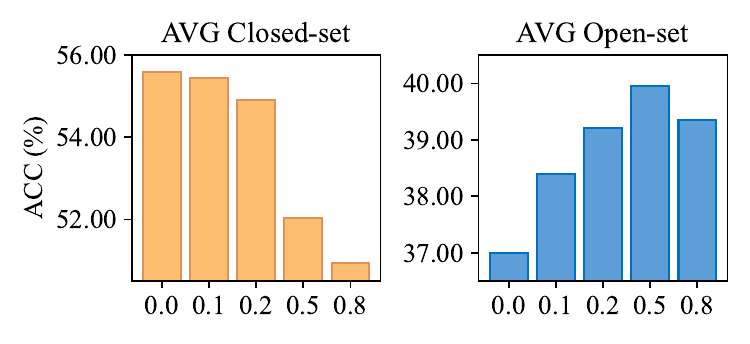}
\vskip -0.15in
\caption{Quantitative analysis of the coefficient $\lambda_{\text{action}}$ for our Action-aware Discrimination loss.
The horizontal axis shows the value of $\lambda_{\text{action}}$.
Introducing our Action-aware Discrimination loss better maintains the open-set accuracy. 
}
\label{fig:exp_con}
\end{minipage}
\hskip 0.2in
\begin{minipage}[t]{0.49\textwidth}
\centering
\includegraphics[width=0.95\linewidth]{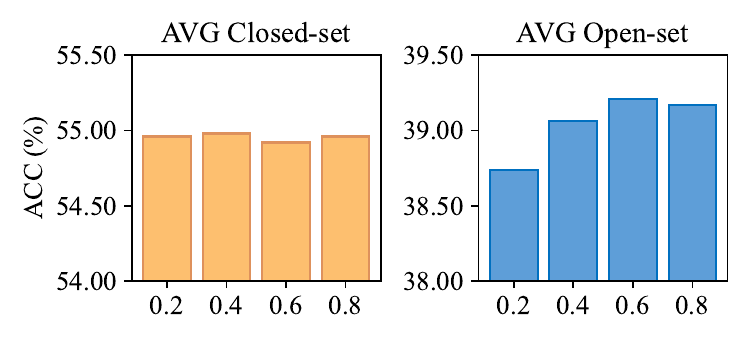}
\vskip -0.15in
\caption{Quantitative analysis of the margin $\delta$ used in our Action-aware Discrimination loss.
The horizontal axis shows the value of $\delta$.
The performance of our Action-aware Discrimination loss is robust to the margin value. 
}
\label{fig:exp_margin}
\end{minipage}
\vskip -0.1in
\end{figure*}

\subsubsection{Effect of Score Ensemble}
During inference, we use both the global video representation $z_{x}$ and elaboration representations $\{\hat{z}_x^{i}\}$ for classification, and introduce an ensemble coefficient $\lambda_{\text{e}}$ to obtain final action classification score.  
In this part, we conduct a quantitative analysis to this ensemble coefficient, and the results are shown in Figure~\ref{fig:exp_ensemble_weight}.
As shown in the figure, our model can obtain better closed-set and open-set recognition performance using different values of ensemble coefficient, compared with our model without Diversified Elaboration Representation Learning (\ie, the red and purple  baselines in the figure). 
By using a smaller ensemble coefficient, our model concentrates more on the open-set recognition since the elaboration score has a larger weight. 
For closed-set recognition, medium values of this coefficient yield the best results. 
By default, we set $\lambda_{\text{e}}=0.4$ according to the best trade-off between closed-set and open-set performance.

\subsection{Quantitative Analysis of Scene-Aware Video-text Alignment}
\subsubsection{Effect of Scene-aware Discrimination}
We conduct a quantitative analysis to the loss coefficient $\lambda_{\text{scene}}$ in our proposed Scene-aware Discrimination loss, and the results on the four test domains are shown in Figure~\ref{fig:exp_dis}.
The results in the figure show that our model consistently obtains better closed-set performance on all test domains with a larger loss weight.
The comprehensive results demonstrate that our proposed Scene-aware Discrimination loss effectively tackles various types of domain gaps by mitigating scene bias in fitting training videos.
However, using only the Scene-aware Discrimination loss will lead to performance drop in open-set generalization, thus we introduce the Action-aware Discrimination loss to address this issue.

\subsubsection{Effect of Action-aware Discrimination}
We first conduct a quantitative analysis to the loss coefficient $\lambda_{\text{action}}$ in our proposed Action-aware Discrimination loss, and the results are shown in Figure~\ref{fig:exp_con}.
As shown in the figure, as the loss weight gradually increases, our model obtains higher open-set accuracy but lower closed-set accuracy.
In addition, when using a very large weight (\eg, $\lambda_{\text{action}}=0.8$), our model cannot obtain higher open-set accuracy. 
This is because our Action-aware Discrimination loss acts as a constraint on the video representation space to compensate our Scene-aware Discrimination loss and is not specifically designed to improve open-set action recognition. 
Also, these results empirically reveal that a good trade-off between the Scene-aware Discrimination and Action-aware Discrimination losses is crucial for generalizable open-vocabulary action recognition.

Second, we conduct a quantitative analysis to the margin $\delta$ in our proposed Action-aware Discrimination loss, and the results are shown in Figure~\ref{fig:exp_margin}.
As shown in the figure, when using different values of $\delta$, our model performs very similarly in terms of cross-domain closed-set accuracy. 
In addition, when using a relatively larger value of $\delta$ (\ie, $0.4\leq \delta \leq 0.8$), our model can obtain promising results in terms of open-set accuracy. 
Overall, our model obtains comparable performance across different settings of $\delta$, which demonstrates that our Action-aware Discrimination loss is robust to the setting of the margin value.

\subsubsection{Effect of Scene Suffixes}
In Figure~\ref{fig:exp_scene}, we show how the suffix number affects the model performance. 
To show the loss effect more clearly, we set the loss coefficient $\lambda_{\text{scene}}=0.5$ in this experiment.
As shown in the figure, our model obtains better closed-set accuracy as the total number of scene suffixes increases.
According to the design of our Scene-aware Discrimination loss, we distinguish videos apart from more scene-encoded text prompts in representation space with more scene suffixes used. 
In this way, videos are less likely to be confused in representation space, leading to stronger abilities for recognizing actions in unseen domains.

\begin{table}[t]
%\small
\fontsize{8.5pt}{14.5pt}\selectfont
\caption{Quantitative analysis of our proposed Scene-aware Discrimination loss $L_{\text{scene}}$ with different types of scene-encoded text prompt templates. 
The reported performance are in terms of the closed-set accuracy. 
}
\label{tab:prompts}
\vskip -0.05in
\centering
\begin{tabular}{c | c | c c c c | c }
\hline Models        & Type   & UCF    & HMDB  & ARID    & NEC-Dr & AVG     \\
\hline
Baseline             & -      & 87.07  & 57.59 & 47.35   & 21.30  & 53.33   \\ 
\hline  
\multirow{3}{*}{+$L_{\text{scene}}$}              
& TypeA  & 89.04  & 56.88 & 47.35   & 19.91  & 53.30   \\
& TypeB  & 87.55  & 57.49 & 49.39   & 20.83  & 53.82   \\
& Ours   & 87.82  & 59.41 & 49.18   & 25.93  & \textbf{55.59}   \\
\hline
\end{tabular}
% \vskip -0.3in
\end{table}

\begin{figure*}[t]
    \begin{minipage}[b]{0.68\textwidth}
        \includegraphics[width=\textwidth]{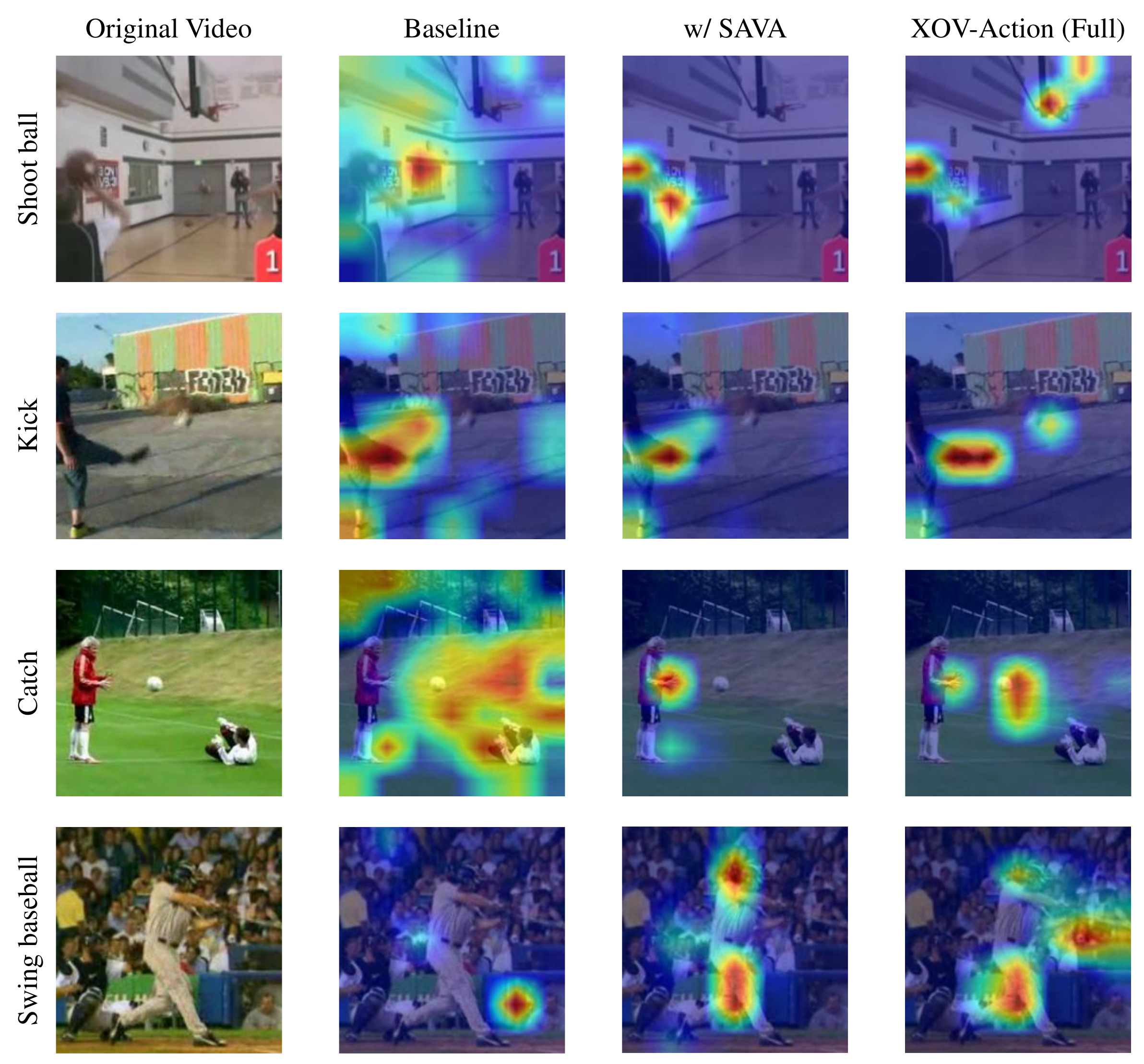}
        \vskip -0.1in
        \caption{
            Qualitative analysis by attention visualization.
            In this figure, we show the original video and the attention visualization of baseline, our model with Scene-Aware Video-text Alignment (w/ SAVA) and our full XOV-Action. 
            Best viewed in color. 
        }
        \label{fig:gradcam}
    \end{minipage}
    \hskip 0.1in
    \begin{minipage}[b]{0.30\textwidth}
        \centering
        \includegraphics[width=\textwidth]{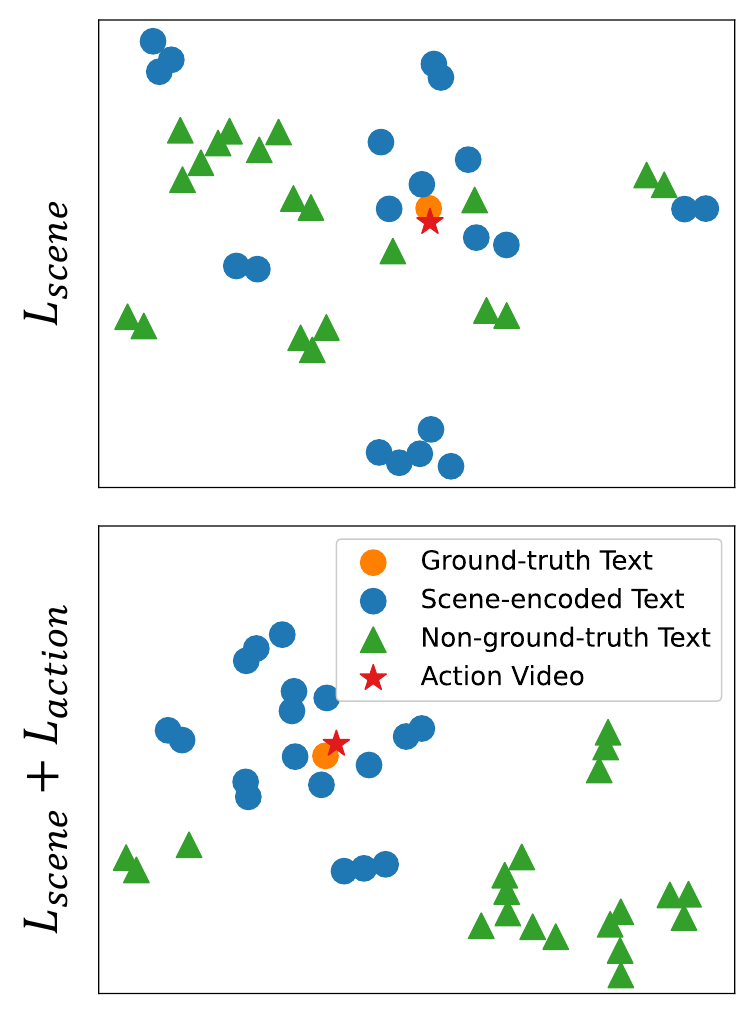}
        \vskip -0.1in
        \caption{
            Qualitative analysis of the Scene-aware Discrimination loss $L_{\text{scene}}$ and Action-aware Discrimination loss $L_{\text{action}}$ by t-SNE. 
            Best viewed in color. 
        }
        \label{fig:tsne}
        \vspace{0.2cm} 
        \includegraphics[width=\textwidth]{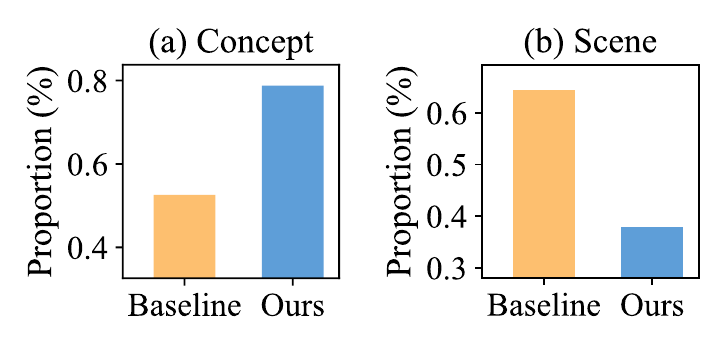}
        \vskip -0.1in
        \caption{
            Quantitative analysis of representation similarity between action videos and (a) concept texts / (b) scene texts.
        }
        \label{fig:exp_similarity}
    \end{minipage}
\vskip -0.1in
\end{figure*}

\subsubsection{Effect of Scene-encoded Text Prompt Templates}
In this part, we conduct a quantitative comparison between our proposed $L_{\text{scene}}$ and its variants that use different types of scene-encoded text prompt templates. 
By default, our proposed $L_{\text{scene}}$ uses the scene-encoded text prompts in the form of ``a video of a person [doing something] [at/on/in the/a scene].'', \eg, for the action ``abseiling'', a scene-encoded text prompt can be ``a video of a person abseiling in the park.''. 
We compare with variants using two different types of scene-encoded text prompts as follows:
(1) TypeA: ``a video of [the/a scene].'', \eg, ``a video of the park.''; 
(2) TypeB: ``a video of a person [at/on/in the/a scene].'', \eg, ``a video of a person in the park.''. 
As shown in Table~\ref{tab:prompts}, the two variants exhibit much lower performance than our proposed $L_{\text{scene}}$ in terms of the average cross-domain closed-set accuracy, which demonstrates that these variants cannot effectively tackle the domain gaps. 
The results imply that encoding the semantic information of the ground-truth action category into scene-encoded text prompts is important to the mitigation of scene bias.

\subsection{In-depth Verification Experiments}
In this subsection, we conduct more verification experiments to demonstrate our model's effectiveness in learning diverse action-related concepts and mitigating scene bias. 

\subsubsection{Qualitative Analysis by Attention Visualization}
In this part, we conduct a detailed qualitative analysis to our model by attention visualization, and the results are shown in Figure~\ref{fig:gradcam}.
As shown in the figure, the baseline pays much attention to video scenes, which will easily cause recognition errors. 
In contrast, by introducing our proposed Scene-Aware Video-text Alignment (denoted by ``w/ SAVA''), our model focuses more on the body parts of action performers rather than scenes, \eg, the hand holding a ball. 
This clearly demonstrate that our Scene-Aware Video-text Alignment can effectively mitigate scene bias and improve action recognition across domains. 
Moreover, by introducing our proposed Diversified Elaboration Representation Learning, our full model can capture more action-related concepts in videos, \eg, the basketball hoop related to the action ``shoot ball''. 
As a result, our model can associate rich learned concepts with novel categories during testing, and improve the open-set action recognition across domains. 

\subsubsection{Qualitative Analysis by Distribution Visualization} 
First, we intuitively show how our proposed Action-aware Discrimination loss $L_{\text{action}}$ affects representation space by t-SNE~\cite{van2008visualizing}, and the results are given in Figure~\ref{fig:tsne}.
As shown by the figure, without $L_{\text{action}}$, our model with only the Scene-aware Discrimination loss may yield higher similarity between a video and some non-ground-truth action name texts than between the video and the scene-encoded text prompts, which would lead to confusion in video classification. 
In contrast, with the Action-aware Discrimination loss, our model produces a much more reasonable representation space: a video is farther from the non-ground-truth action texts than from the scene-encoded text prompts. 
This experiment intuitively demonstrates the rationale of our proposed Action-aware Discrimination loss.

\subsubsection{Quantitative Analysis of Scenes and Concepts} 
To demonstrate our model's effectiveness in learning diverse action-related concepts and mitigating scene bias, we conduct an in-depth quantitative analysis to the representation similarity between action videos and specific types of texts in our XOV-Action model. 
First, we analyze the representation similarity between videos and concept texts. 
We ask GPT-4 to generate a large pool of words or phrases that describes concepts related to human actions, \eg, football, goalposts, kickers and legs are some common concepts related to the action ``kicking field goal''.
Then, we extract the video representations of training samples, and compute the representation similarity between these video representations and the representations of the generated concept texts. 
We calculate the proportion of concept texts that have high video-text similarity scores with at least one video sample (we also set the threshold as 0.3), and the statistical results are shown in Figure~\ref{fig:exp_similarity} (a). 
As shown in the figure, compared with the baseline, the video representations of our proposed XOV-Action have high similarity with more concept texts, indicating that our model encodes more action-related concept information. 
This demonstrates that our proposed Diversified Elaboration Representation Learning can effectively capture rich action-related concepts, which boosts the understanding of open-set action categories. 

Second, we focus on analyzing the representation similarity between videos and scene texts. 
We ask GPT-4 to generate a large pool of words or phrases about action scenarios, \eg, park, house, court. 
Similar to the analysis of the scene texts, we calculate the proportion of scene texts that have video-text similarity scores greater than 0.3 with at least one video sample. 
Figure~\ref{fig:exp_similarity} (b) shows the statistics of the baseline and our XOV-Action.
As shown in the figure, compared with the baseline without our Scene-Aware Video-text Alignment, our XOV-Action learns video representations that are generally more dissimilar to scene text representations, which demonstrates our model's effectiveness in mitigating the scene bias.

\section{Conclusion}
This work concentrated on a valuable but underexplored task, namely generalizable open-vocabulary action recognition.
To address this task, we proposed a novel model named XOV-Action, aiming to overcome two critical challenges of this task: (1) understanding novel action concepts of open-set categories, and (2) mitigating the scenario discrepancy between the training and test domains. 
First, XOV-Action proposed to capture rich action-related concepts by learning diversified elaboration representations, enhancing its recognition of open-set action categories. 
Second, XOV-Action proposed to mitigate the scene bias by learning scene-agnostic video representations, thereby improving the generalization across video domains. 
In addition, we contributed a new benchmark named XOVABench, which provided a comprehensive way to evaluate models across various types of domain gaps and different action categories.  
Extensive quantitative and qualitative experiments showed that our proposed XOV-Action can effectively improve both closed-set and open-set action recognition performance across domains. 
We believe that our work will serve as a catalyst for further advancement in the field of robust video understanding, and we hope it will inspire future innovative solutions for the vision-language understanding field.

\section*{Acknowledgments}
{
The authors would like to thank Bing Zhao and Zhi-Wei Xia for their support in writing and experiments.
This work was supported by the New Generation Artificial Intelligence-National Science and Technology Major Project (2025ZD0123100). 
Additionally, this work was partially supported by NSFC  (92470202), National Key Research and Development Program of China (2023YFA1008503), Guangdong NSF Project (No. 2023B1515040025), Guangdong Key Research and Development Program (No. 2024B0101040004, No. 2025B0909020002).
}

\section*{Appendix}

\setcounter{section}{0}
\renewcommand{\thesection}{A.\arabic{section}}

\renewcommand{\thefigure}{A\arabic{figure}} 
\renewcommand{\thetable}{A\arabic{table}}   
\setcounter{figure}{0} 
\setcounter{table}{0}  
\renewcommand{\theequation}{A\arabic{equation}}
\setcounter{equation}{0}

\section{Text and Prompt Formulation and Usage}
Our proposed XOV-Action model utilizes different types of text or prompts to facilitate the learning of generalizable open-vocabulary action recognition models. 
Specifically, these text and prompts can be divided into three types:

\noindent\textbf{1) Text prompt of action name}: This kind of text prompt is in the form of ``\texttt{a video of a person [doing something].}''. For example, for the action category ``long jump'', the text prompt of this action is ``\texttt{a video of a person long jump.}''.    

\noindent\textbf{2) Textual descriptions of action}: For each action category, we adopt $M$ automatically generated textual descriptions for training (generated by GPT-4~\cite{openai2023gpt4}). For example, for the action category ``long jump'', these textual descriptions can be:
  \begin{itemize}
    \item \texttt{a person is seen accelerating on a track, taking a leap, and landing in a distant sand pit.}
    \item \texttt{a person is sprinting before jumping as far as possible into a sand-filled pit.}
    \item \texttt{...}
    \item \texttt{a person propels their body forward in a horizontal leap, landing in a sand pit.}
  \end{itemize}
  As shown above, these descriptions concisely capture the key visual aspects of the action, and these descriptions are used in our proposed Diversified Elaboration Representation Learning. 

\noindent\textbf{3) Scene-encoded text prompt}: For each action category, we introduce $N$ scene-encoded text prompts for training. Each scene-encoded text prompt consists of an action and a scene. For example, for the action category ``long jump'', these scene-encoded text prompts can be:
  \begin{itemize}
    \item \texttt{a video of a person long jumping in the park.}
    \item \texttt{...}
    \item \texttt{a video of a person long jumping on the street.}
    \item \texttt{a video of a person long jumping in the kitchen.}
  \end{itemize}
 scene-encoded text prompts are used in our proposed Scene-Aware Video-text Alignment.

\section{Analysis of Description Usage}
In this part, we conduct additional experiments to analyze the usage of action descriptions.
First, we quantitatively analyze the effects of description sources by using different large language models to generate textual descriptions.
As shown in Table~\ref{tab:supp-desc}, our proposed XOV-Action model achieves comparable performance when using descriptions generated by different large language models, and consistently outperforms the previous state-of-the-art (\ie, Open-VCLIP).
These results highlight the effectiveness of our model and its robustness to the choice of description source.

\IEEEpubidadjcol

\begin{table}[h]
\vskip -0.1in
%\small
\fontsize{7.5pt}{11.5pt}\selectfont
\caption{
Quantitative analysis of XOV-Action using action descriptions generated by different large language models. 
In this table, we report the average closed-set, open-set and overall performance on XOVABench. 
}
\label{tab:supp-desc}
\vskip -0.05in
\centering
{
\setlength{\tabcolsep}{3pt}
\begin{tabular}{c | c | c c c }
\hline Models        & Description Sources     & Closed & Open & All       \\
\hline
Open-VCLIP~\cite{DBLP:conf/iclr/GuLKC22}           & -
& 53.00           & 39.08     
& 46.95\\  
\hline
\multirow{6}{*}{XOV-Action}  
& DeepSeek-V3.2~\cite{deepseekai2025deepseekv32pushingfrontieropen}
& 54.63           & 42.03     
& 49.22 \\    
& gemma-3-27b-it~\cite{gemmateam2025gemma3technicalreport}
& 54.57           & 41.96     
& 49.16 \\     
& Qwen2.5-32B-Instruct~\cite{qwen2025qwen25technicalreport}
& 54.66           & 42.29     
& 49.39 \\   
& Qwen3-30B-A3B-Instruct~\cite{yang2025qwen3technicalreport}
& 54.88           & 42.06     
& 49.36 \\   
& Qwen3-4B-Instruct~\cite{yang2025qwen3technicalreport}
& 54.50           & 42.22     
& 49.26 \\   
& GPT-4~\cite{openai2023gpt4} (default)                           
& 55.96           & 41.79     
& 49.80\\
\hline
\end{tabular}
}
\vskip -0.05in
\end{table}

Second, we conduct experiments to demonstrate our model' robustness to prompt phrasing and model temperature during the LLM-based description generation process. 
In this experiment, we use the open-source Qwen3-30B-A3B-Instruct-2507~\cite{yang2025qwen3technicalreport} as the LLM generator. 
Specifically, we adopt three different prompts in this experiments, and try three different model temperatures, \ie, 0.2, 0.7 and 1.0. 
The default model temperature is 0.7.
We adopt three different prompts in this experiments, which are shown in Prompt~\ref{pr:prompt-v1}, Prompt~\ref{pr:prompt-v2} and Prompt~\ref{pr:prompt-v3}. 
PromptV1 is the default prompt we used, and PromptV2 and PromptV3 are automatically generated by LLMs with minimal manual refinement.
In each prompt, the ``\texttt{\{\}}'' will be filled by the action label text when generating descriptions for a specific action category. 
As shown in Table~\ref{tab:supp-desc-llmpara}, our XOV-Action obtains comparable performance across different settings, with the overall performance gap between models being less than 0.8\%. 
This demonstrates our model's robustness to prompt phrasing and model temperature for LLMs. 
\begin{promptbox}{}{prompt-v1}
\textbf{Prompt V1:} Generate 10 descriptions for \{\} human action in a video, each description should be concise, that is less than 40 words. These descriptions are required to be diverse, but should be distinguishable to describe \{\} human action. Each description consists of only one sentence, starts with 'A person', and ends up with a full stop '.'. Please return these 10 sentence in a JSON list form. The list would only contains 10 strings, without any other outputs.
\end{promptbox}
\begin{promptbox}{}{prompt-v2}
\textbf{Prompt V2:} You are given an action label: ``\{\}''.\\
Write exactly 10 distinct one-sentence descriptions of a video showing this action.\\ \\
Constraints: \\
- Each sentence must be <= 40 words. \\
- Each sentence must start with ``A person'' and end with a period ``.''. \\
- Describe only observable human action and context; do not speculate about invisible causes or intentions. \\
- The 10 sentences must be diverse in wording and details but clearly depict the same action ``\{\}''. \\ \\
Output format: \\
Return ONLY a JSON array of 10 strings, with no extra text, no numbering, and no Markdown.
\end{promptbox}
\begin{promptbox}{}{prompt-v3}
\textbf{Prompt V3:} Task: Generate concise, diverse captions for a single human action class in a video. \\ \\
Action class: ``\{\}'' \\
Requirements: \\
1) Produce 10 captions total. \\
2) One caption per line conceptually, but your final output must be a JSON list of 10 strings. \\
3) Every caption: \\
  - begins with "A person" \\
  - is a single sentence \\
  - is under 40 words \\
  - ends with "." \\
4) Captions should vary in phrasing, scene context, and objects involved, while still describing the same action ``\{\}''. \\ \\
Return only the JSON list (10 strings). Do not add explanations or any other tokens.
\end{promptbox}

\begin{table}[t]
\fontsize{7.5pt}{11.5pt}\selectfont
\caption{
Results of our XOV-Action model using different prompts and different model temperatures during the LLM-based description generalization process. 
We adopt the open-source Qwen3-30B-A3B-Instruct-2507~\cite{yang2025qwen3technicalreport} as the LLM generator in this experiment. 
The default model temperature is 0.7. 
In this table, we report the average closed-set, open-set and overall performance on XOVABench. 
}
\label{tab:supp-desc-llmpara}
\vskip -0.05in
\centering
\begin{tabular}{c | c | c c c }
\hline Models        & Description Sources     & Closed & Open & All       \\
\hline                       
Open-VCLIP~\cite{DBLP:conf/iclr/GuLKC22}           & -
& 53.00           & 39.08     & 46.95 \\     
\hline                       
\multirow{5}{*}{XOV-Action}
& Prompt V1 (default)
& 54.88           & 42.06     & 49.36 \\   
& Prompt V2
& 55.21           & 42.12     & 49.50 \\   
& Prompt V3
& 54.60           & 41.03     & 48.79 \\   
& Prompt V1 \& Temp 0.2
& 55.45           & 41.90     & 49.62 \\     
& Prompt V1 \& Temp 1.0
& 54.75           & 42.17     & 49.34 \\  
\hline
\end{tabular}
\end{table}

Third, we conduct additional experiments using human-written descriptions and templated LLM-generated descriptions. 
In this experiment, we also use the open-source Qwen3-30B-A3B-Instruct-2507~\cite{yang2025qwen3technicalreport} as the LLM generator. 
\begin{itemize}
  \item For human-written descriptions, we use action definition from Chen et al.~\cite{DBLP:conf/iccv/ChenH21}, which are refined by human labor. As there is only one action definition for each action category, we replace one Qwen-generated description with the action definition for training. As shown by ``Prompt V1 w/ Human'' in Table~\ref{tab:supp-desc-humanandtemp}, such a model variant achieves very similar performance to the default version, which demonstrates that human-written descriptions can be reliable for model training. In practice, we recommend prioritizing LLMs for description generation, since this avoids costly human labor and the capability is already relatively mature. 
  \item Additionally, we conduct another experiment where we constrain the LLM to generate descriptions based on a fixed set of 10 templates. The specific prompt is shown in Prompt~\ref{pr:prompt-v4}, \ie, Prompt V4. As shown in Table~\ref{tab:supp-desc-humanandtemp}, this template-based prompting degrades the model's open-set performance. We attribute this to the reduced description diversity for each action category, which limits the model's ability to learn a broader range of action-related concepts. Therefore, in practical applications, we do not recommend using templated LLM prompts.
  \begin{promptbox}{}{prompt-v4}
  \textbf{Prompt V4:} Generate 10 descriptions for \{\} human action in a video, each description should be concise, that is less than 40 words. These descriptions are required to be diverse, but should be distinguishable to describe \{\} human action. Each description consists of only one sentence, starts with 'A person', and ends up with a full stop '.'. Here are some templates for generating descriptions, and please a different template each time: \\
  - "A person is \{\{motion\}\} with \{\{object\}\}, \{\{motion\}\}." \\
  - "A person performs \{\{motion\}\} on \{\{object\}\}, \{\{motion\}\}." \\
  - "A person is performing \{\{motion\}\} using \{\{object\}\}, \{\{motion\}\}." \\
  - "A person carries out \{\{motion\}\} involving \{\{object\}\}, \{\{motion\}\}." \\
  - "A person is engaged in \{\{motion\}\} with \{\{object\}\}, \{\{motion\}\}." \\
  - "A person continues \{\{motion\}\} with \{\{object\}\}, \{\{motion\}\}." \\
  - "A person starts \{\{motion\}\} with \{\{object\}\}, \{\{motion\}\}." \\
  - "A person focuses on \{\{motion\}\} with \{\{object\}\}, \{\{motion\}\}." \\ 
  - "A person completes \{\{motion\}\} with \{\{object\}\}, \{\{motion\}\}." \\
  - "A person \{\{motion\}\} with \{\{object\}\}, \{\{motion\}\}." \\
  Please fill in the blanks \{\{motion\}\} or \{\{object\}\} adaptively according to the given action category. Please return these 10 sentence in a JSON list form. The list should only contains 10 strings, without any other outputs.
  \end{promptbox}
\end{itemize}

\begin{table}[t]
% \vskip -0.1in
\fontsize{7.5pt}{11.5pt}\selectfont
\caption{
Results of our XOV-Action model using human-written descriptions and templated LLM-generated descriptions. 
In this table, we report the average closed-set, open-set and overall performance on XOVABench. 
}
\label{tab:supp-desc-humanandtemp}
\vskip -0.05in
\centering
\begin{tabular}{c | c | c c c }
\hline Models        & Description Sources     & Closed & Open & All       \\
\hline                       
Open-VCLIP~\cite{DBLP:conf/iclr/GuLKC22}           & -
& 53.00           & 39.08     & 46.95 \\     
\hline                       
\multirow{3}{*}{XOV-Action}
& Prompt V1 (default)
& 54.88           & 42.06     & 49.36 \\   
& Prompt V1 w/ Human
& 54.95           & 41.87     & 49.40 \\   
& Prompt V4 (Templated)
& 55.16           & 40.82     & 48.93 \\ 
\hline
\end{tabular}
\vskip -0.05in
\end{table}

Finally, we conduct additional experiments to analyze the robustness to description redundancy and hallucination, and the results demonstrate that our model is robust to a small amount of redundancy or hallucination. 
\begin{itemize}
    \item For description redundancy, we intentionally inject exact duplicate descriptions into the description set.
    Specifically, for each action category, we deliberately replace several of the ten descriptions with \textit{an identical description}, making some descriptions fully redundant (\ie, semantically identical) to each other. 
    As shown in Table~\ref{tab:supp-desc-redundant}, open-set performance gradually decreases as more duplicate descriptions are used, since the description set becomes less diverse. 
    However, the overall degradation is modest and our model remains effective compared with previous state-of-the-art models, demonstrating that our approach is robust to mild description redundancy. 
    \item For description hallucination, we intentionally inject exact incorrect descriptions into the description set. 
    Specifically, for each action category, we deliberately replace several of the ten descriptions with \textit{incorrect descriptions}, \ie, descriptions from other action categories. 
    As shown in Table~\ref{tab:supp-desc-wrong}, a small amount of hallucination does not have a significant impact on our model's performance, \eg, one or two incorrect descriptions per category. 
    This is attributed to the design of our elaboration selection strategy, which can filter out some irrelevant descriptions for training videos. 
    As more incorrect descriptions are introduced, performance drops significantly, especially on open-set categories, since these incorrect descriptions are more likely to provide misleading supervision signals. 
    These results demonstrate our model' robustness to a small amount of description hallucination. 
\end{itemize}

\begin{table}[t]
\fontsize{7.5pt}{11.5pt}\selectfont
\caption{
Results of our XOV-Action model using duplicate textual descriptions.
The column ``\# of Repeated Descs'' indicates the number of repeated descriptions among the introduced descriptions for each action category. 
In this table, we report the average closed-set, open-set and overall performance on XOVABench. 
}
\label{tab:supp-desc-redundant}
\vskip -0.05in
\centering
\begin{tabular}{c | c | c c c }
\hline Models        & \# of Repeated Descs & Closed    & Open  & All  \\
\hline                       
Open-VCLIP~\cite{DBLP:conf/iclr/GuLKC22}          & -
& 53.00           & 39.08     & 46.95 \\     
\hline                       
\multirow{4}{*}{XOV-Action}
& 0 (default)
& 55.96           & 41.79     & 49.80 \\  
& 1
& 56.12           & 41.67     & 49.88 \\    
& 3
& 55.88           & 41.31     & 49.55 \\    
& 5
& 55.73           & 41.08     & 49.28 \\   
\hline
\end{tabular}
\end{table}

\begin{table}[t]
\fontsize{7.5pt}{11.5pt}\selectfont
\caption{
Results of our XOV-Action model using some hallucinated textual descriptions.
The column ``\# of Hallucinated Descs'' indicates the number of hallucinated descriptions among the introduced descriptions for each action category. 
In this table, we report the average closed-set, open-set and overall performance on XOVABench. 
}
\label{tab:supp-desc-wrong}
\vskip -0.05in
\centering
\begin{tabular}{c | c | c c c }
\hline Models        & \# of Hallucinated Descs & Closed    & Open  & All  \\
\hline                       
Open-VCLIP~\cite{DBLP:conf/iclr/GuLKC22}           & -
& 53.00           & 39.08     & 46.95 \\     
\hline                       
\multirow{4}{*}{XOV-Action}
& 0 (default)
& 55.96           & 41.79     & 49.80 \\  
& 1
& 56.14           & 41.75     & 49.89 \\    
& 2
& 55.72           & 41.96     & 49.80 \\    
& 4
& 54.30           & 39.41     & 47.99 \\     
\hline
\end{tabular}
\vskip -0.05in
\end{table}

Overall, these above results demonstrate that our model is robust to the adopted action descriptions.

\begin{table}[t]
\vskip -0.0in
\fontsize{8.5pt}{12.5pt}\selectfont
\caption{
Quantitative analysis of the elaboration selection strategy. 
In this table, we report the average closed-set and open-set performance on XOVABench. 
}
\label{tab:elaboration_strategies}
\vskip -0.05in
\centering
\begin{tabular}{c | c c }
\hline  
Strategy      & ~Closed-set~    & ~Open-set~   \\
\hline
~top-$\hat{C}$ highest similarity (default)~         
& 55.96           & 41.79      \\
random            
& 53.83           & 41.41      \\
top-$\hat{C}$ lowest similarity              
& 52.57           & 41.16      \\
\hline
\end{tabular}
\end{table}

\begin{figure}[!t]
\centering
\includegraphics[width=1.0\linewidth]{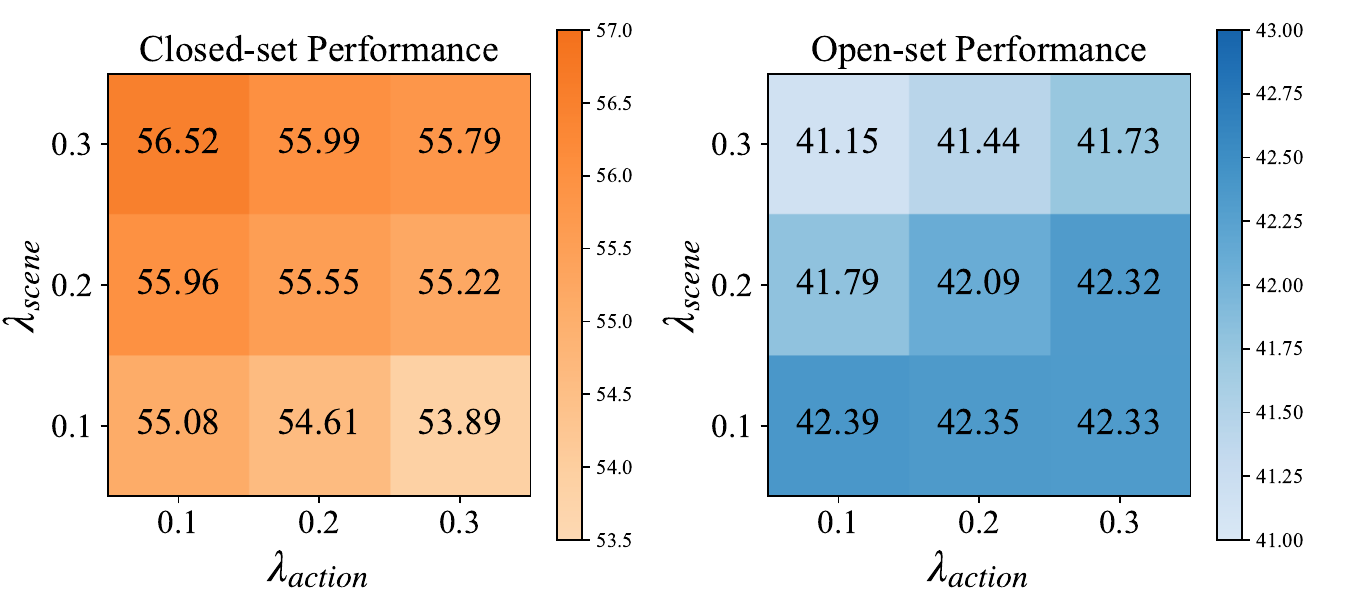}
\vskip -0.12in
\caption{
Quantitative analysis of the joint effects of $\lambda_{\text{scene}}$ and $\lambda_{\text{action}}$. 
In this figure, we report the average closed-set and open-set performance on XOVABench. 
}
\label{fig:hyper-negpos}
\end{figure}

\begin{figure}[!t]
\vskip -0.1in
\centering
\includegraphics[width=1.0\linewidth]{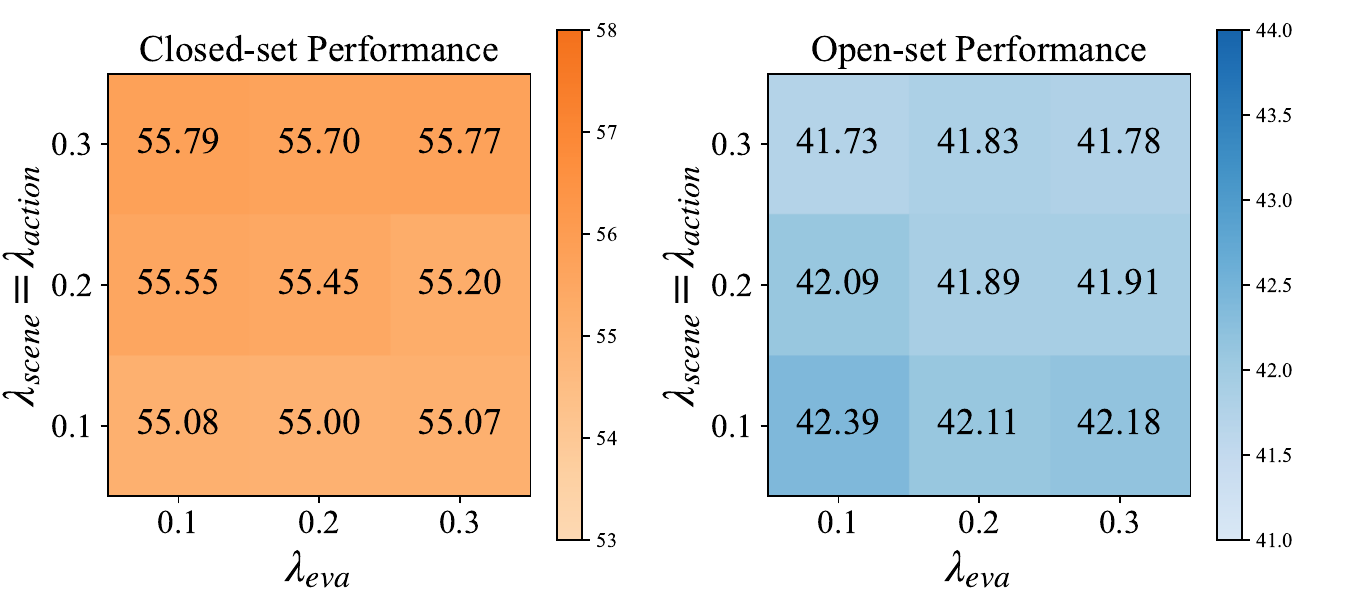}
\vskip -0.12in
\caption{
Quantitative analysis of the joint effects of $\lambda_{\text{scene}}$, $\lambda_{\text{action}}$ and $\lambda_{\text{eva}}$. 
In this experiment, we set $\lambda_{\text{scene}}=\lambda_{\text{action}}$, and we report the average closed-set and open-set performance on XOVABench. 
}
\label{fig:hyper-negposeva}
\end{figure}

\section{Effect of Elaboration Selection Strategy}
In this part, we quantitatively analyze the effects of elaboration selection strategies on our model. 
As shown in Table~\ref{tab:elaboration_strategies}, random selection reduces open-set performance, since $\hat{C}$ descriptions are randomly selected irrespective of their similarity. 
In this case, closed-set performance also decreases since noisy or irrelevant descriptions are likely to be selected, leading to unreliable video-text alignment during training. 
Furthermore, selecting the top-$\hat{C}$ lowest similarity obtains lower performance in both closed-set and open-set recognition. 
This is because the selected descriptions are the most irrelevant to the video content, which introduces more severe noise than random selection. 
Overall, these results demonstrate the effectiveness of our employed strategy.

\section{Analysis of Joint Hyperparameter Effects}
In this part, we conduct additional experiments to analyze the joint effects of the three loss coefficients, namely $\lambda_{\text{scene}}$, $\lambda_{\text{action}}$ and $\lambda_{\text{eva}}$. 
Figure~\ref{fig:hyper-negpos} shows the joint effects of $\lambda_{\text{scene}}$ and $\lambda_{\text{action}}$. As shown in the figure, closed-set performance generally improves as $\lambda_{\text{scene}}$ increases. This is because our Scene-aware Discrimination loss $L_{\text{scene}}$ encourages the video encoder to downweight the attention on scene information in videos, thereby improving generalization in unseen domains. However, introducing $L_{\text{scene}}$ reduces open-set performance, as open-set performance drops as $\lambda_{\text{scene}}$ increases. Therefore, we propose the Action-aware Discrimination loss $L_{\text{action}}$ to constrain video representation learning and alleviate the degradation of open-set performance. As shown in Figure~\ref{fig:hyper-negpos}, increasing $\lambda_{\text{action}}$ yields relatively higher open-set performance. 

Figure~\ref{fig:hyper-negposeva} shows the joint effects of $\lambda_{\text{scene}}$, $\lambda_{\text{action}}$ and $\lambda_{\text{eva}}$, where we set $\lambda_{\text{scene}}=\lambda_{\text{action}}$. As shown in the figure, increasing $\lambda_{\text{scene}}$ and $\lambda_{\text{action}}$ improves closed-set performance but leads to a decrease in open-set performance, suggesting that these coefficients should be balanced in practice. Additionally, varying $\lambda_{\text{eva}}$ within the range $[0.1, 0.3]$ leads to only minor performance changes, demonstrating that our method is relatively insensitive to this coefficient. Note that removing the loss $L_{\text{eva}}$ causes a significant performance drop, \ie, 55.96\%~$\to$~55.20\% in closed-set recognition and 41.79\%~$\to$~40.66\% in open-set recognition, which verifies the effectiveness of $L_{\text{eva}}$.

\begin{table}[t]
\vskip -0.0in
\fontsize{8.5pt}{12.5pt}\selectfont
\caption{
Additional ablation study on more loss function combinations.
In this table, we report the average closed-set and open-set performance for each model variant on XOVABench. 
}
\label{tab:morecombination}
\vskip -0.05in
\centering
{\renewcommand{\arraystretch}{1.15}
  \begin{tabular}{c|ccc|cc}
    \hline
    Models  & $L_{\text{scene}}$ & $L_{\text{action}}$ & $L_{\text{eva}}$~ & Closed-set    & Open-set \\
    \hline
    w/o $L_{\text{scene}}$  
    & \xmark     & \checkmark & \checkmark & 53.85 & 42.41 \\
    w/o $L_{\text{action}}$    
    & \checkmark & \xmark     & \checkmark & 56.16 & 40.99 \\
    w/o $L_{\text{eva}}$    
    % & \checkmark & \checkmark & \xmark     & 55.06 & 40.56 \\
    & \checkmark & \checkmark & \xmark     & 55.20 & 40.66 \\
    \hline
    Full
    & \checkmark & \checkmark & \checkmark & 55.96 & 41.79 \\
    \hline
  \end{tabular}
}
\vskip -0.1in
\end{table}

\section{More Combinations of Losses}
In this part, we conducted an additional ablation study by including more combinations of loss functions. 
Specifically, our XOV-Action model includes three loss contributions, namely Scene-aware Discrimination $L_{\text{scene}}$, Action-aware Discrimination $L_{\text{action}}$, and Elaborative Video-text Alignment $L_{\text{eva}}$. 
We start from the full model and individually remove each loss term. 
With the exception of the modified loss functions, we keep the network architecture and inference strategy identical in this experiment. 
The results are summarized in Table~\ref{tab:morecombination}, and we discuss the results as follows:
\begin{itemize}    
  \item Removing the Scene-aware Discrimination loss $L_{\text{scene}}$ from our model leads to a significant performance drop in cross-domain closed-set action recognition, as shown by ``w/o $L_{\text{scene}}$''. 
  This is because $L_{\text{scene}}$ is designed to mitigate domain gaps by downweighting the attention on scene information in videos. 
  Consequently, its removal reduces the closed-set performance in unseen domains. 
  
  \item Removing the Action-aware Discrimination loss $L_{\text{action}}$ from our model leads to a significant performance drop in open-set action recognition, as shown by ``w/o $L_{\text{action}}$''. 
  This is because $L_{\text{action}}$ is designed to alleviate the degradation of cross-domain open-set performance caused by the introduction of $L_{\text{scene}}$.
  Thus, the removal of $L_{\text{action}}$ degrades open-set performance. 
  
  \item Removing the Elaborative Video-text Alignment $L_{\text{eva}}$ loss from our model leads to a performance drop in both closed-set and open-set action recognition, with the drop in open-set recognition being more significant, as shown by ``w/o $L_{\text{eva}}$''.   
  This is because our Elaborative Video-text Alignment is designed to learn more action-related concept information and facilitate the understanding of novel concepts in open-set categories.
  Therefore, removing $L_{\text{eva}}$ hinders the model's ability to generalize to diverse open-set categories.
\end{itemize}    
Overall, these additional results demonstrate the effectiveness of our loss contributions and are consistent with our previous findings.

\section{Generalizability in Egocentric Scenarios}
To further demonstrate the effectiveness of our model, we construct an additional egocentric benchmark based on EPIC-Kitchen~\cite{DBLP:conf/eccv/DamenDFFFKMMPPW18} and evaluate model performance in egocentric scenarios using this benchmark. 
Our evaluation shows that our proposed XOV-Action can also outperform previous state-of-the-art models in egocentric scenarios. 
Specifically, the additional egocentric benchmark, which we refer to as EPIC-XOV, is a first-person view benchmark for generalizable open-vocabulary action recognition. 
Its videos are captured in kitchen environments from an egocentric viewpoint, and it comprises 34 closed-set categories and 63 open-set categories. 
The comparison results between our model and previous state-of-the-art models are summarized in Table~\ref{tab:epic}. 
As shown in the table, all models exhibit relatively limited performance on this benchmark, particularly on the open-set categories. 
This can be largely attributed to the significant domain gap between the training dataset and EPIC-XOV. 
Specifically, the training data mainly consist of third-person daily activity videos, where visual cues are human-centric, whereas EPIC-XOV contains egocentric videos that predominantly capture interactions among hands, objects, and the surrounding environment without seeing action performers themselves. 
As a result, degraded performance is expected under such a challenging cross-domain setting.
Despite this difficulty, our method still achieves the best performance on this benchmark, demonstrating its effectiveness and generalizability.
We leave further improving the transferability of action recognition models from third-person videos to egocentric videos as an important direction for future work.

\begin{table}[t]
\vskip -0.0in
\caption{
Comparison with Kinetics400-trained open-vocabulary action recognition models on EPIC-XOV egocentric benchmark. 
In this table, we report the average closed-set, open-set and overall performance on EPIC-XOV. 
}
\label{tab:epic}
\vskip -0.05in
\centering
{\renewcommand{\arraystretch}{1.15}
\begin{tabular}{c | c c c }
\hline Models        & Closed-set    & Open-set  & ~~All~~  \\
\hline                       
{ActionCLIP}~\cite{DBLP:journals/corr/abs-2109-08472}
& 17.64      & 0.56      & 9.71  \\                      
{ViFiCLIP}~\cite{DBLP:conf/cvpr/RasheedK0KK23}
& 14.48      & 0.71      & 8.08  \\ 
{OpenVCLIP}~\cite{DBLP:conf/icml/WengYLWJ23}
& 20.57      & 0.16      & 11.09 \\    
{FROSTER}~\cite{DBLP:journals/corr/abs-2402-03241}
& 19.01      & 0.16      & 10.26 \\ 
\hline               
\rowcolor{pink!20}        
~XOV-Action (Ours)~           
& \textbf{27.12}           &  \textbf{1.02}     & \textbf{15.00} \\   
\hline
\end{tabular}
}
\vskip -0.1in
\end{table}

% \section*{References}
{
   \small
   \bibliographystyle{unsrt}
   \bibliography{clipref}
}

\vfill

\end{document}